%% file: main.tex
\documentclass[10pt,twocolumn,letterpaper]{article}

\usepackage[pagenumbers]{cvpr} %

\input{preamble}

\definecolor{cvprblue}{rgb}{0.21,0.49,0.74}
\usepackage[pagebackref,breaklinks,colorlinks,citecolor=cvprblue]{hyperref}

\def\paperID{xxx} %
\def\confName{xxx}
\def\confYear{xxx}

\title{Segment Anything Meets Point Tracking}

\author{
 Frano Raji\v{c}$^{1,3}$\hspace{0.35cm}Lei Ke$^{1,2}$\hspace{0.35cm}Yu-Wing Tai$^2$\hspace{0.35cm}Chi-Keung Tang$^2$\hspace{0.35cm}Martin Danelljan$^1$\hspace{0.35cm}Fisher Yu$^1$ \\
 $^1$ETH Z{\"u}rich\hspace{1.5cm}$^2$HKUST\hspace{1.5cm}$^3$EPFL\hspace{1.5cm} \\
 }
 
\begin{document}
\maketitle

\begin{strip}
    \vspace{-0.4in}
    \centering
    \includegraphics[width=\linewidth]{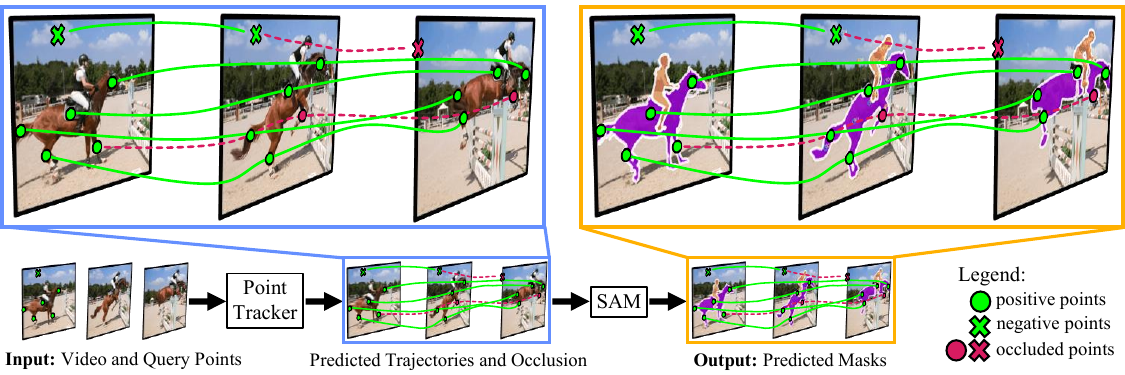}
    \captionof{figure}{
        {Segment Anything Meets Point Tracking (SAM-PT).}
        SAM-PT is a \textit{point-centric} method that utilizes sparse point propagation for interactive video segmentation, enabling easier interaction and faster annotation. We extend SAM~\cite{kirillov2023segment} with long-term point trackers to effectively operate on videos in a \textit{zero-shot} manner. SAM-PT takes user clicks as ``query points'' which either denote the target object (positive points) or designate non-target segments (negative points). The points are tracked throughout the video using point trackers that propagate the query points to all video frames, producing trajectory predictions and occlusion scores. SAM is subsequently prompted with the non-occluded points in the trajectories as to output a segmentation mask for each video frame independently. The propagated points can be further edited for accurate segmentation and tracking.
        \label{fig:pull-figure}
    }
    \vspace{-0.1in}
\end{strip}

\begin{abstract}

The Segment Anything Model (SAM) has established itself as a powerful zero-shot image segmentation model, enabled by efficient point-centric annotation and prompt-based models. While click and brush interactions are both well explored in interactive image segmentation, the existing methods on videos focus on mask annotation and propagation. This paper presents SAM-PT, a novel method for point-centric interactive video segmentation, empowered by SAM and long-term point tracking. SAM-PT leverages robust and sparse point selection and propagation techniques for mask generation. Compared to traditional object-centric mask propagation strategies, we uniquely use point propagation to exploit local structure information agnostic to object semantics. We highlight the merits of point-based tracking through direct evaluation on the zero-shot open-world Unidentified Video Objects (UVO) benchmark. 
Our experiments on popular video object segmentation and multi-object segmentation tracking benchmarks, including DAVIS, YouTube-VOS, and BDD100K, suggest that a point-based segmentation tracker yields better zero-shot performance and efficient interactions. We release our code that integrates different point trackers and video segmentation benchmarks at \url{https://github.com/SysCV/sam-pt}.
    
\end{abstract}

\section{Introduction}
\label{sec:intro}

Object segmentation and tracking in videos are central pillars for a myriad of applications, including autonomous driving, robotics, and video editing. Despite significant progress made in the past few years with deep neural networks~\cite{yang2022decoupling,cheng2022xmem,wu2021seqformer,cheng2021mask2former}, we still need to rely on expensive labels for model supervision to achieve high accuracies. Therefore, many efforts have been made on interactive video segmentation to accelerate the data labeling process and benefit artistic video editing. Those methods are usually evaluated on a simplified semi-supervised segmentation setup as a proxy for full interactive video segmentation.

The prevailing methods~\cite{cheng2022xmem,cheng2021mask2former} in semi-supervised Video Object Segmentation (VOS) and Video Instance Segmentation (VIS) exhibit performance gaps when dealing with unseen data, particularly in a zero-shot setting, \textit{i.e.}, when these models are transferred to video domains they have not been trained or that encompass object categories falling outside of the training distribution.

Further, generalizable models usually require large amounts of training data. The existing interactive video segmentation methods assume the mask of an object is given on the first frame of the testing video. While getting accurate mask is laborious, recent works~\cite{kirillov2023segment,cheng2022pointly} on training foundation image segmentation models show that point-based annotation in combination with mask editing tools is a scalable approach to label exceedingly large amounts of data. Despite its success on images in terms of labeling efficiency and accuracy, point-centric interactive segmentation has received scant attention in the video domains.

In this paper, we aim to achieve both domain generalizability and labeling efficiency for interactive video segmentation. Our insights are two-fold. First, foundation models in image segmentation are available, such as Segment Anything Model (SAM)~\cite{kirillov2023segment}. SAM, trained on 11 million images and 1 billion object masks, has impressive zero-shot generalization capabilities. The model also supports point prompts as additional inputs for interactive image segmentation and produces high-quality masks. Second, we witnessed significant recent progress in point tracking~\cite{doersch2023tapvid,harley2022particle,zheng2023point,doersch2023tapir,wang2023omnimotion,karaev2023cotracker}. Those tracking methods, once trained, can propagate points across video frames on diverse domains.

Therefore, we introduce SAM-PT (Segment Anything Meets Point Tracking), depicted in~\cref{fig:pull-figure}. This is the first method to utilize sparse point tracking combined with SAM for video segmentation, offering a new perspective on solving the problem. Instead of employing object-centric dense feature matching or mask propagation, we propose a point-centric approach that capitalizes on tracking points using rich local structure information embedded in videos. It only requires sparse points annotation to denote the target object in the first frame and provides better generalization to unseen objects. This approach also helps preserve the inherent flexibility of SAM while extending its capabilities effectively to video segmentation. Similar to the data annotation process in SAM, our point-centric approach can be potentially integrated with the existing mask-based approaches in real-world applications.

SAM-PT prompts SAM with sparse point trajectories predicted using state-of-the-art point trackers, such as CoTracker~\cite{karaev2023cotracker}, harnessing their versatility for video segmentation. We identified that initializing points to track using K-Medoids cluster centers from a mask label was the strategy most compatible with prompting SAM. Tracking both positive and negative points enables the clear delineation of target objects from their background. To further refine the output masks, we propose multiple mask decoding passes that integrate both types of points. In addition, we devised a point reinitialization strategy that increases tracking accuracy over time. This approach involves discarding points that have become unreliable or occluded, and adding points from object parts or segments that become visible in later frames, such as when the object rotates.

We evaluate SAM-PT on multiple setups including semi-supervised, open-world, and fully interactive video segmentation. Our method achieves stronger performance than existing zero-shot methods
by up to $5.0\%$ on DAVIS, $2.0\%$ on YouTube-VOS, and $7.3\%$ on BDD100K, while also surpassing a fully-supervised VIS method~\cite{yang2023track} on UVO by $6.7$ points. We also set up a new benchmark for interactive point-based video segmentation to simulate the process of manually labeling the whole video. In this setup, SAM-PT significantly reduces annotation effort, approaching the performance of fully supervised approaches and underscoring its practicality. This comes without the need for any video segmentation data during training, underscoring the robustness and adaptability of our approach, and indicating its potential to enhance progress in video segmentation tasks, particularly in zero-shot scenarios.

\section{Related Work}
\label{sec:related}

\noindent {\bf Point Tracking for Video Segmentation.}\quad
Classical feature extraction and tracking methods such as Lucas-Kanade~\cite{lucas1981iterative}, Tomasi-Kanade~\cite{tomasi1991detection}, Shi-Tomasi~\cite{shi1994good}, SIFT~\cite{lowe2004distinctive}, and SURF~\cite{bay2008speeded}, as well as newer methods such as LIFT~\cite{yi2016lift}, SuperPoint~\cite{detone2018superpoint}, and SuperGlue~\cite{sarlin2020superglue}, have all demonstrated proficiency in identifying or tracking sparse features and establishing long-range correspondences. Nonetheless, these techniques often falter in dynamic, non-rigid environments. While flow-based approaches such as RAFT~\cite{teed2020raft} offer improvements, they too struggle with maintaining long-term point accuracy due to error accumulation and occlusions. Addressing these shortcomings, recent innovations such as
PIPS~\cite{harley2022particle}, PIPS++~\cite{zheng2023point}, OmniMotion~\cite{wang2023omnimotion}, TAPIR~\cite{doersch2023tapir}, and the state-of-the-art CoTracker~\cite{karaev2023cotracker}, optimize for robust long-term trajectories and effectively manage occlusions. Our work is unique in applying these methods to guide image segmentation models for video segmentation tasks.

\noindent \textbf{Segment and Track Anything Models.}\quad
SAM~\cite{kirillov2023segment} is a foundation model for image segmentation that showcases impressive zero-shot capabilities. Its extension, HQ-SAM~\cite{ke2023segment}, improves mask quality for complex objects but is not designed for video tasks. TAM~\cite{yang2023track} and SAM-Track~\cite{cheng2023segment} attempt to extend SAM to video segmentation by integrating the state-of-the-art fully-supervised XMem~\cite{cheng2022xmem} and DeAOT~\cite{yang2022deaot} mask trackers, respectively, yet they lack in zero-shot scenarios.

\noindent \textbf{Zero-Shot VOS / VIS.}\quad
Generalist models such as Painter~\cite{wang2023images} apply visual prompting to various tasks but demonstrate limited performance in video segmentation. On the other hand, SegGPT~\cite{wang2023seggpt} also uses visual prompting and competes closely with our method on some datasets. Other approaches, such as STC~\cite{jabri2020space} and DINO~\cite{caron2021emerging}, perform VOS through feature matching. Our approach distinguishes itself by taking the point-centric approach to enhance performance on VOS benchmarks in a zero-shot setting.

\noindent \textbf{Interactive VOS.}\quad
Interactive VOS has shifted from labor-intensive manual annotations to more user-friendly interaction methods, such as scribbles, clicks, and drawings, enabling rapid and intuitive video editing~\cite{wang2005interactiveVideoCutout, oh2019fastInteractive, miao2020memoryAggregationInteractive, Caelles_arXiv_2019, Yuk2020IVOSGlobalLocal}. Among these, MiVOS~\cite{cheng2021mivos} stands out for its modular design that decouples mask generation from propagation, effectively incorporating user interactions from diverse interaction modalities. Unlike MiVOS and other fully-supervised methods, SAM-PT is the first to use point propagation instead of mask propagation and thus operates effectively in zero-shot settings. Our interactive point-based video segmentation study emphasizes the simplicity and efficacy of point interactions and differs from common scribble-based benchmarking~\cite{Caelles_arXiv_2018} or in-distribution user studies~\cite{cheng2021mivos}.

\section{Method}
\label{sec:method}

We propose SAM-PT for addressing video segmentation tasks in a zero-shot setting. SAM-PT combines the strengths of the Segment Anything Model (SAM), a foundation model for image segmentation, and prominent point trackers, such as PIPS~\cite{harley2022particle} and CoTracker~\cite{karaev2023cotracker}, to enable interactive tracking of anything in videos. \cref{method:sam-preliminaries} briefly describes the background knowledge about SAM. \cref{method:samp} then introduces our SAM-PT method with its four constituent steps. Finally, \cref{method:samp-vs-traditional} analyzes and highlights the method's novelty as the first point-centric interactive video segmentation method compared to existing works.

\subsection{Preliminaries: SAM}
\label{method:sam-preliminaries}

Whereas in computer vision ``zero-shot (learning)'' usually refers to the study of generalization to unseen object categories in image classification~\cite{lampert2009learning}, we follow prior work~\cite{kirillov2023segment, radford2021learning} and rather employ the term in a broader sense and explore generalization to unseen datasets.

The Segment Anything Model (SAM)~\cite{kirillov2023segment} is a novel vision foundation model designed for promptable image segmentation. SAM is trained on the large-scale SA-1B dataset, which contains 11 million images and over 1 billion masks. SA-1B has 400 times more masks than any prior segmentation dataset. This extensive training set facilitates SAM's impressive zero-shot generalization capabilities. SAM has showcased its ability to produce high-quality masks from a single foreground point and has demonstrated robust generalization capacity on a variety of downstream tasks under a zero-shot transfer protocol using prompt engineering. These tasks include, but are not limited to, edge detection, object proposal generation, and instance segmentation.

SAM comprises three main components: an image encoder, a flexible prompt encoder, and a fast mask decoder. The image encoder is a Vision Transformer (ViT) backbone and processes high-resolution $1024 \times1024$ images to generate an image embedding of $64 \times 64$ spatial size. The prompt encoder takes sparse prompts as input, including points, boxes, and text, or dense prompts such as masks, and translates these prompts into $c$-dimensional tokens. The lightweight mask decoder then integrates the image and prompt embeddings to predict segmentation masks in real-time, allowing SAM to adapt to diverse prompts with minimal computational overhead.

\subsection{Ours: SAM-PT}
\label{method:samp}

While SAM shows impressive capabilities in image segmentation, it is inherently limited in handling video segmentation tasks. Our Segment Anything Meets Point Tracking (SAM-PT) approach effectively extends SAM to videos, offering robust video segmentation without requiring training on any video segmentation data.

\begin{figure*}
    \centering
    \includegraphics[width=\linewidth]{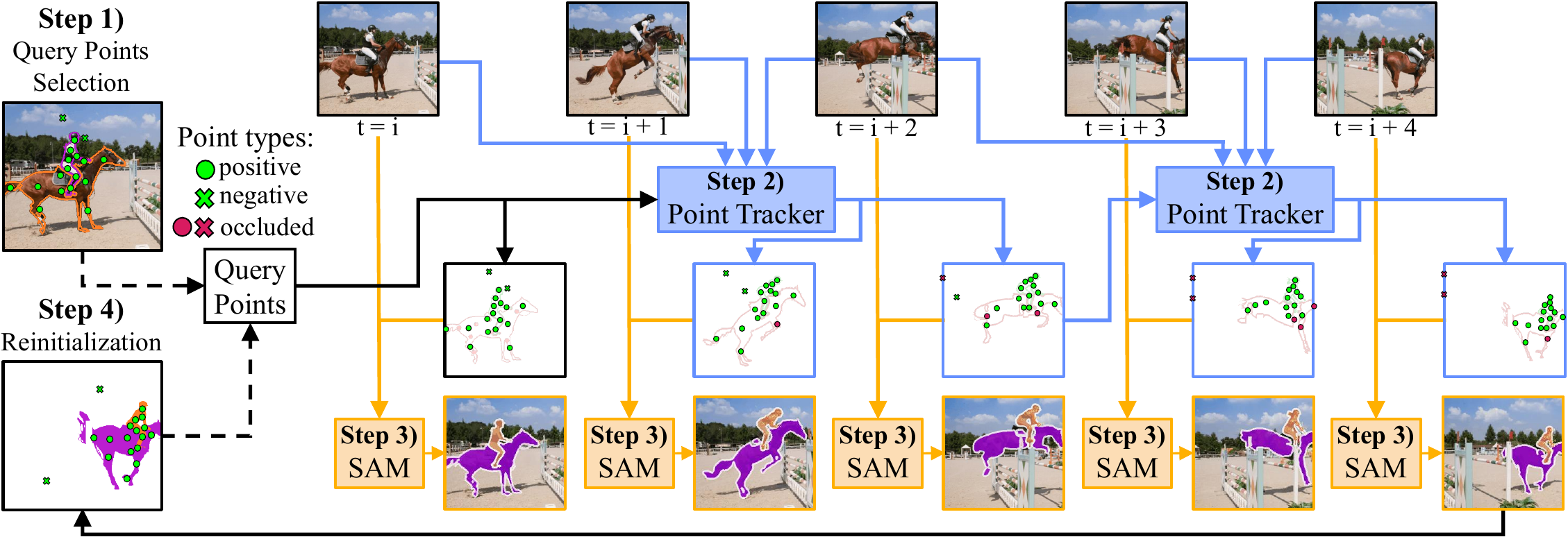}
    \caption{
    {Segment Anything Meets Point Tracking (SAM-PT) overview.} The essence of SAM-PT is to extend image segmentation foundation models to effectively operate on videos.
    SAM-PT has four steps: $\mathbf{1}$\textbf{) Query Points Selection.} It starts with first-frame query points which denote the target object (positive points) or designate non-target segments (negative points). These points are provided by the user or derived from a ground truth mask. $\mathbf{2}$\textbf{) Point Tracking.} Initiated with the query points, our approach leverages point trackers to propagate the points across video frames, predicting point trajectories and occlusion scores. $\mathbf{3}$\textbf{) Segmentation.} The trajectories are then used to prompt the Segment Anything Model (SAM) and output per-frame mask predictions. $\mathbf{4}$\textbf{) Point Tracking Reinitialization.} Optionally, the predicted masks are used to reinitialize the query points and restart the process when reaching a prediction horizon $h$.
    Reinitialization helps by getting rid of unreliable points and adding points to object segments that become visible in later frames.
    }
    \label{fig:system-figure}
    \vspace{-0.1in}
\end{figure*}

SAM-PT is illustrated in \cref{fig:system-figure} and is primarily composed of four steps: \textbf{1)} selecting query points for the first frame; \textbf{2)} propagating these points to all video frames using point trackers; \textbf{3)} using SAM to generate per-frame segmentation masks based on the propagated points;~\textbf{4)} optionally reinitializing the process by sampling query points from the predicted masks. We next elaborate on these four steps.

\begin{figure}
    \centering
    \includegraphics[width=\linewidth]{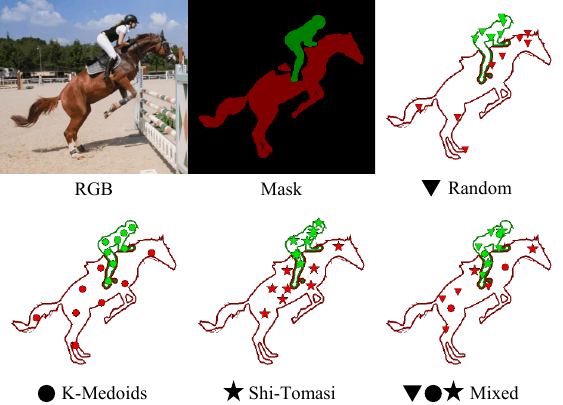}
    \caption{
    {Positive Point Sampling.} For an image paired with either a ground truth or predicted segmentation mask, positive points are sampled from within the mask area using one of the following point sampling methods: Random, K-Medoids~\cite{park2009kmedoids}, Shi-Tomasi~\cite{shi1994good}, or Mixed. Notably, Random Sampling and K-Medoids Sampling only require the segmentation mask for input, not the corresponding input image. For negative points, we always use Mixed Sampling on the target object's background mask.
    }
    \label{fig:point-sampling-methods}
    \vspace{-0.2in}
\end{figure}

\noindent \textbf{1) Query Points Selection.}\quad
The process begins with defining query points in the first video frame, which either denote the target object (positive points) or designate the background and non-target objects (negative points). Users can manually and interactively provide query points, or they may be derived from a ground truth mask. For example, in the case of semi-supervised video object segmentation, the ground truth mask is provided for the first frame where the object appears. We derive the query points from ground truth masks using different point sampling techniques by considering their geometrical locations or feature dissimilarities, as depicted in \cref{fig:point-sampling-methods}. These sampling techniques are:

\begin{itemize}
    \item \textbf{Random Sampling:} An intuitive approach where query points are randomly selected from the ground truth mask.
    \item \textbf{K-Medoids Sampling:} This technique takes the cluster centers of K-Medoids clustering~\cite{park2009kmedoids} as query points to ensure good coverage of different parts of the object and robustness to noise and outliers.
    \item \textbf{Shi-Tomasi Sampling:} This method extracts Shi-Tomasi corner points from the image under the mask as they have been shown to be good features to track~\cite{shi1994good}.
    \item \textbf{Mixed Sampling:} A hybrid method combining the above techniques since it might benefit from the unique strengths of each.
\end{itemize}

While each method contributes distinct characteristics that influence the model's performance, our ablation study reveals that K-Medoids sampling yields the best results with good coverage of various segments of the complete object. Shi-Tomasi sampling follows closely, indicating their respective strengths in this context. The selection and arrangement of these points considerably affect the overall video segmentation performance, thus determining the optimal method is crucial.

\noindent \textbf{2) Point Tracking.}\quad
Initiated with the query points, we employ robust point trackers to propagate the points across all frames in the video, resulting in point trajectories and occlusion scores. We adopt point trackers such as PIPS~\cite{harley2022particle} and the state-of-the-art CoTracker~\cite{karaev2023cotracker} to propagate the points as they show moderate robustness toward long-term tracking challenges such as object occlusion and re-appearance. Long-term point trackers are also shown more effective than methods such as chained optical flow propagation or first-frame correspondences in our experiments.

\begin{figure}
    \centering
    \includegraphics[width=\linewidth]{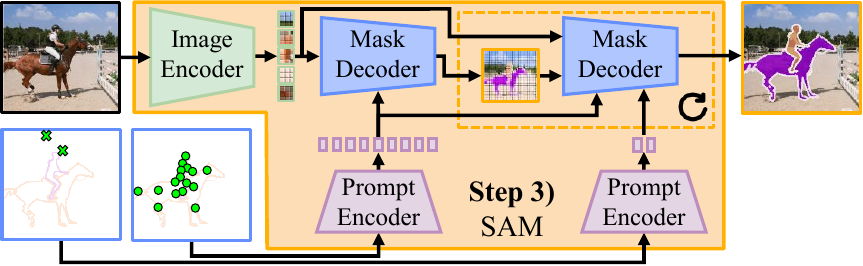}
    \caption{
    {Interacting with SAM in SAM-PT.}
    In the first pass, SAM is prompted exclusively with positive points to define the object's initial localization. In the second pass, both positive and negative points along with the previous mask prediction are fed to the same mask decoder for further mask refinement. The negative points remove segments from the background and neighboring objects and notably help in cases when the point tracker mistakenly predicts positive points off the target object. The second pass is repeated iteratively to get a refined segmentation mask.
    }
    \label{fig:sam}
    \vspace{-0.2in}
\end{figure}

\noindent \textbf{3) Segmentation.}\quad
In the predicted trajectories, the non-occluded points serve as indicators of where the target object is throughout the video. This allows us to use the non-occluded points to prompt SAM, as illustrated in \cref{fig:sam}, and leverage its inherent generalization ability to output per-frame segmentation mask predictions. Unlike conventional tracking methods that require training or fine-tuning on video segmentation data, our approach excels in zero-shot video segmentation tasks.

We combine positive and negative points by calling SAM in two passes. In the initial pass, we prompt SAM exclusively with positive points to define the object's initial localization. Subsequently, in the second pass, we prompt SAM with both positive and negative points along with the previous mask prediction. Negative points provide a more nuanced distinction between the object and the background and help by removing wrongly segmented areas.

Lastly, we execute a variable number of mask refinement iterations by repeating the second pass. This utilizes SAM's capacity to refine vague masks into more precise ones. Based on our ablation study, this step notably improves video object segmentation performance.

\noindent \textbf{4) Point Tracking Reinitialization.}\quad
We optionally execute a reinitialization of the query points using the predicted masks once a prediction horizon of $h=8$ frames is reached. Upon reaching this horizon, we have $h$ predicted masks and will take the last one to sample new points. At this stage, all previous points are discarded and substituted with the newly sampled points. Following this, steps 1) through 4) are repeated with the new points, starting from the horizon timestep where reinitialization occurs. The steps are iteratively executed until the entire video is processed. The reinitialization process serves to enhance tracking accuracy over time by discarding unreliable or occluded points while incorporating points from object segments that become visible later in the video. Other reinitialization variants are discussed in our Supplementary Material. %

\subsection{SAM-PT vs. Object-centric Mask Propagation}
\label{method:samp-vs-traditional}

With sparse point tracking combined with prompting SAM, SAM-PT distinguishes itself from traditional video segmentation methods that depend on dense object mask propagation, as noted in \cref{tab:mask-propagation-methods}. To propagate the first-frame GT label to the remaining video frames, traditional techniques commonly use feature matching with masks cached to a mask memory~\cite{yang2023track,cheng2023segment,cheng2022xmem,yang2022decoupling}, frame-by-frame feature matching~\cite{jabri2020space,caron2021emerging},
optical flow~\cite{yang2021self},
and, recently, in-context visual prompting~\cite{wang2023images,wang2023seggpt}. In contrast, SAM-PT introduces a unique approach to video object segmentation, employing the robust combination of point tracking with SAM, which is inherently designed to operate on sparse point prompts.

The point propagation strategy of SAM-PT offers several advantages over traditional object-centric tracking methods. First, point propagation exploits local structure context that is agnostic to global object semantics.
This enhances our model's capability for zero-shot generalization, an advantage that, coupled with SAM's inherent generalization power, allows for tracking diverse objects in diverse environments, such as on the UVO benchmark.
Second, SAM-PT allows for a more compact object representation with sparse points, capturing enough information to characterize the object's segments/parts effectively. Finally, the use of points is naturally compatible with SAM, an image segmentation foundation model trained to operate on sparse point prompts, offering an integrated solution that aligns well with the intrinsic capacities of the underlying model. 

Comparing SAM-PT with conventional methods in \cref{tab:mask-propagation-methods}, SAM-PT emerges as superior or comparable to methods that refrain from utilizing video segmentation data during training. However, there is a performance gap that exists between such methods and those that leverage video segmentation training data in the same domain, such as XMem~\cite{cheng2022xmem} or DeAOT~\cite{yang2022decoupling}. Further, the potential of our model extends beyond video object segmentation to other tasks, such as Video Instance Segmentation (VIS), thanks to the inherent flexibility of our point propagation strategy.

\begin{table}[!t]
\footnotesize

    \caption{
    Comparative analysis of semi-supervised Video Object Segmentation methods. Our approach, SAM-PT, introduces \textit{sparse point propagation}, a compact mask representation that uses local structure information agnostic to object semantics. It outperforms other non-video-data-dependent methods, achieving top $\mathcal{J\&F}$ scores on DAVIS $2016$ and $2017$, and the highest $\mathcal{G}$ score on YouTube-VOS $2018$. The comparison considers the reliance on video mask data during training, zero-shot learning setting, initial frame label requirements, and label propagation techniques used.
 }
        \vspace{-0.07in}
	\centering
        \resizebox{\columnwidth}{!}{
		\begin{tabular}{lcccccccc}
            \toprule
            \textbf{Method}
            & \makecell[c]{\textbf{Video}\\ \textbf{Mask}}
            & \makecell[c]{\textbf{Zero-}\\ \textbf{Shot}}
            & \makecell[c]{\textbf{Frame}\\ \textbf{Init.}}
            & \textbf{Propagation}
            & \makecell[c]{\textbf{DAVIS}\\$\mathbf{2016}$}
            & \makecell[c]{\textbf{DAVIS}\\$\mathbf{2017}$}
            & \makecell[c]{\textbf{YTVOS}\\$\mathbf{2018}$}
            \\
            \midrule
            SiamMask~\cite{wang2019fast}         & \ding{51} & \ding{55} & Box          & Feature Correlation & \gr{$69.8$} & \gr{$56.4$} & \gr{-} \\
            QMRA~\cite{lin2021query}             & \ding{51} & \ding{55} & Box          & Feature Correlation & \gr{$85.9$} & \gr{$71.9$} & \gr{-} \\
            TAM~\cite{yang2023track}             & \ding{51} & \ding{55} & Points       & Feature Matching    & \gr{$88.4$} & \gr{-}      & \gr{-} \\
            SAM-Track~\cite{cheng2023segment}    & \ding{51} & \ding{55} & Points       & Feature Matching    & \gr{$92.0$} & \gr{-}      & \gr{-} \\
            DEVA~\cite{cheng2023tracking}        & \ding{51} & \ding{55} & Mask         & Feature Matching    & \gr{-}      & \gr{$87.6$} & \gr{-} \\
            XMem~\cite{cheng2022xmem}            & \ding{51} & \ding{55} & Mask         &  Feature Matching   & \gr{$92.0$} & \gr{$\textbf{87.7}$} & \gr{$86.1$} \\
            DeAOT~\cite{yang2022decoupling}      & \ding{51} & \ding{55} & Mask         &  Feature Matching   & \gr{$\mathbf{92.9}$} & \gr{$86.2$} & \gr{$\mathbf{86.2}$} \\ \midrule
            Painter~\cite{wang2023images}        & \ding{55} & \ding{51} & Mask         &  Mask Prompting     & - & $34.6$ & $24.1$ \\
            STC~\cite{jabri2020space}            & \ding{55} & \ding{51} & Mask         &  Feature Matching   & - & $67.6$ & - \\
            DINO~\cite{caron2021emerging}        & \ding{55} & \ding{51} & Mask         &  Feature Matching   & - & $71.4$ & - \\
            SegGPT~\cite{wang2023seggpt}         & \ding{55} & \ding{51} & Mask         & Mask Prompting           & $82.3$      & $75.6$ & $74.7$  \\
            \rowcolor{cyan!16} SAM-PT~(ours) & \ding{55} & \ding{51} & Points       & \textbf{Points Prompting}          & $\mathbf{84.3}$ & $\mathbf{79.4}$ & $\mathbf{76.2}$ \\
            \bottomrule
        \end{tabular}
        }
	\label{tab:mask-propagation-methods}
 \vspace{-0.1in}
 
\end{table}

\section{Experiments}
\label{sec:exp}

\subsection{Datasets}
\label{sec:exp-datasets}
We evaluate our method on four VOS datasets: DAVIS $2016$, DAVIS $2017$~\cite{ponttuset2018davis}, YouTube-VOS $2018$~\cite{xu2018youtubevos}, and MOSE $2023$~\cite{ding2023mose}. DAVIS $2017$ is also used in our interactive point-based video segmentation study. We additionally devise a VOS dataset from BDD100K~\cite{bdd100k}. For VIS, We evaluate our method on the class-agnostic dense video instance segmentation task of the UVO v$1.0$~\cite{wang2021unidentified} dataset. UVO v$1.0$ is a VIS dataset aiming for open-world segmentation, where objects of any category, including those unseen in training, are identified and segmented.

\subsection{Implementation Details}
\label{sec:exp-implementation}

\noindent \textbf{Training Data.}\quad For our experiments, we use pre-trained checkpoints provided by the respective authors for the point trackers (PIPS~\cite{sand2008particle}, CoTracker~\cite{karaev2023cotracker}, etc.) and SAM. PIPS and CoTracker have been trained exclusively on synthetic data, PIPS on FlyingThings++~\cite{harley2022particle}
and CoTracker on TAP-Vid-Kubric~\cite{doersch2023tapvid}.
SAM has been trained on the large-scale SA-1B dataset, the largest image segmentation dataset to date.
HQ-SAM is further trained on the HQ-Seg-44k~\cite{ke2023segment}. Noteworthy, none of these datasets contain video segmentation data, nor do they intersect with any datasets we use for evaluation, situating our model within a zero-shot setting.

\vspace{0.5mm}
\noindent \textbf{Interactive Point-Based Video Segmentation.}\quad
To assess the interactive capabilities of SAM-PT, we simulate user refinement of video segmentation results through point additions and removals. We compare three methods: a non-tracking approach using SAM alone, an online method making one pass through the video with a target IoU quality, and an offline method employing multiple passes that progressively aim at higher IoU quality. These methods are detailed in the Supplementary Material, which includes pseudocode for each simulation variant.

\subsection{Ablation Study}
\label{sec:exp-ablation}

Our ablation experiments on the DAVIS $2017$ validation subset assessed different aspects of SAM-PT's design. Despite the valuable insights, we acknowledge the dataset's limited scope in representing diverse and complex segmentation challenges, such as occlusions and varying environmental conditions, may constrain the generalizability of our ablation's findings. Future investigations could benefit from a more varied validation set, potentially sourced from the YouTube-VOS $2018$ training dataset, to enhance robustness.

Our findings, detailed in \cref{table:ablation-point-trackers}, underscore SAM-PT's adaptability across leading long-term point trackers -- PIPS~\cite{harley2022particle}, TAPIR~\cite{doersch2023tapir}, and notably CoTracker~\cite{karaev2023cotracker}, which excelled due to its precise point tracking and reliable occlusion predictions. In contrast, PIPS++~\cite{zheng2023point} lagged despite being a more recent iteration of PIPS\cite{harley2022particle}, due to the lack of occlusion prediction which is important for the effective use of point tracking for segmentation. TapNet~\cite{doersch2023tapvid} struggled due to less effective temporal consistency and high-resolution inputs. Traditional methods such as SuperGlue~\cite{sarlin2020superglue} and RAFT~\cite{teed2020raft}, which, although proficient in their respective domains, either struggle with the dynamic and deformable aspects of video scenes or cannot handle occlusion, highlighting the specialized efficacy of long-term trackers in the video segmentation landscape.

\input{tables/ablation-point-trackers}

In \cref{table:ablation-pips}, we tested SAM-PT with various settings using PIPS as the tracker. We found that using eight positive points per object instead of just one improved our scores significantly by $33.4$ points because one point often wasn't enough for unambiguously prompting SAM. Selecting points with K-Medoids was slightly better than random and matched Shi-Tomasi, giving a boost of $1.8$ points. Incorporating negative points besides positive points helped when trackers made mistakes, such as losing track of an object, by improving scores by another $1.8$ points. Adding iterative refinement smoothed out mask quality and fixed some errors, adding another $2.2$ points to our performance. We tried filtering out unreliable points with patch similarity, but this did not work well as it ended up removing too many points. Finally, although reinitializing points did not help significantly in the initial tests, it did show benefits on other datasets such as MOSE and UVO, helping us recover from tracker errors by discarding incorrect and adding fresh points as well as detecting that the object has disappeared and the tracking should be halted.

\input{tables/ablation-pips}

Extended ablation experiments and discussions can be found in the Supplementary Material, including complete ablation results for PIPS and CoTracker, the choice of the SAM backbone, and SAM's lightweight variants that reveal trade-offs between performance and inference speed.

\subsection{Video Object Segmentation}
\label{sec:exp-results-vos}

\noindent \textbf{Performance Overview.}\quad
Our SAM-PT method, utilizing HQ-SAM and CoTracker, sets a new standard in zero-shot video object segmentation on the DAVIS $2017$ dataset with a mean $\mathcal{J\&F}$ score of $79.4$, outperforming SegGPT's $75.6$, DINO's $71.4$, and Painter's $34.6$ as shown in \cref{table:vos-davis-2017-val}. On the easier DAVIS $2016$ validation set, our method achieves $84.3$, surpassing SegGPT's $82.3$, showcasing the strength of our approach even in less complex scenarios, as detailed in the Supplementary Material.

\input{tables/davis-2017-valid}

For the YouTube-VOS $2018$ validation set, we achieve the highest performance among zero-shot methods with $76.2$ against SegGPT's $74.7$ and Painter's $24.1$, indicating robust generalizability across various video segmentation benchmarks (\cref{table:vos-ytvos-2018-val}).
\input{tables/ytvos-2018-valid}
In the semi-supervised VOS on BDD100K's validation set, our method outperforms SegGPT for non-transient objects but also surpasses the fully-supervised XMem across nearly all object visibility durations. The detailed breakdown is provided in \cref{table:vos-bdd100k-val}.

On the MOSE $2023$ validation set, our performance remains competitive with SegGPT, with exact figures available in the Supplementary Material.

\begin{figure*}
    \centering
        \begin{subfigure}{\linewidth}
        \centering
        \includegraphics[width=\linewidth]{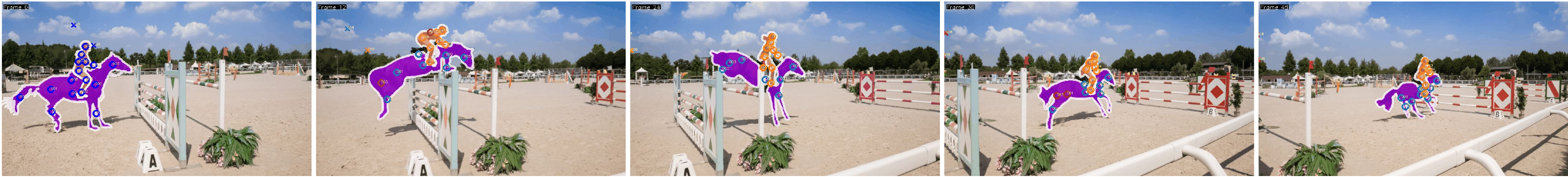}
        \includegraphics[width=\linewidth]{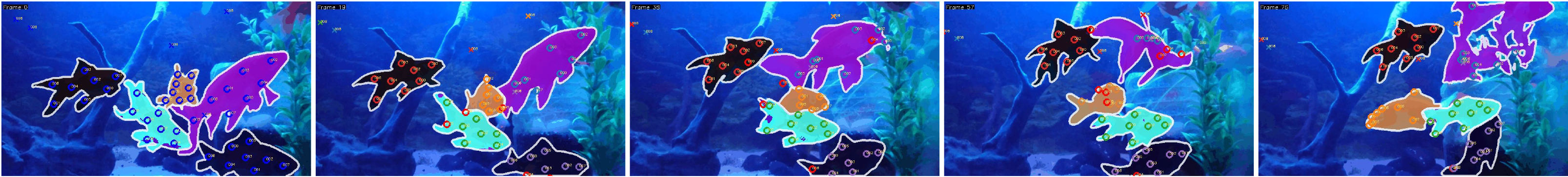}
        \caption{Successful segmentation cases for SAM-PT using 8 positive and 1 negative point.}
        \label{davis-good-cases-samp}
    \end{subfigure}
    \centering
        \begin{subfigure}{\linewidth}
        \centering
        \vspace{0.12cm}
        \includegraphics[width=\linewidth]{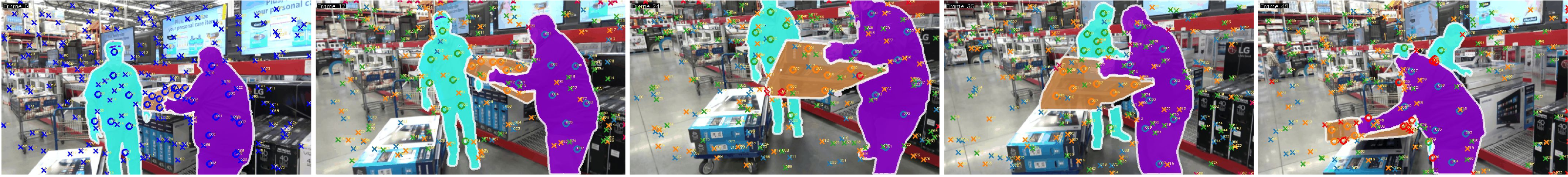}
        \caption{Successful segmentation cases for SAM-PT with 8 positive and 72 negative points and reinitialization enabled.}
        \label{davis-good-cases-samp-reinit}
    \end{subfigure}
    \centering
        \begin{subfigure}{\linewidth}
        \centering
        \vspace{0.12cm}
        \includegraphics[width=\linewidth]{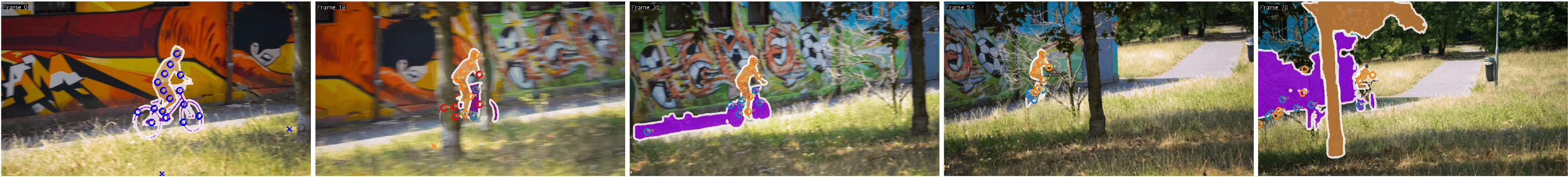}
        \caption{Failure cases for SAM-PT where challenges such as occlusions and thin object structures lead to tracking errors.}
        \label{davis-failure-cases-samp}
    \end{subfigure}
    \vspace{-0.28in}
    \caption{Visualization of SAM-PT on DAVIS 2017~\cite{ponttuset2018davis}. The method shows its capability to segment and track objects using the initial masks from the first frame, with circles denoting positive points and crosses negative points. Red symbols indicate occlusion prediction.
    }
    \label{fig:qualitative-results}
\end{figure*}

\begin{figure*}
    \centering
    \includegraphics[width=\linewidth]{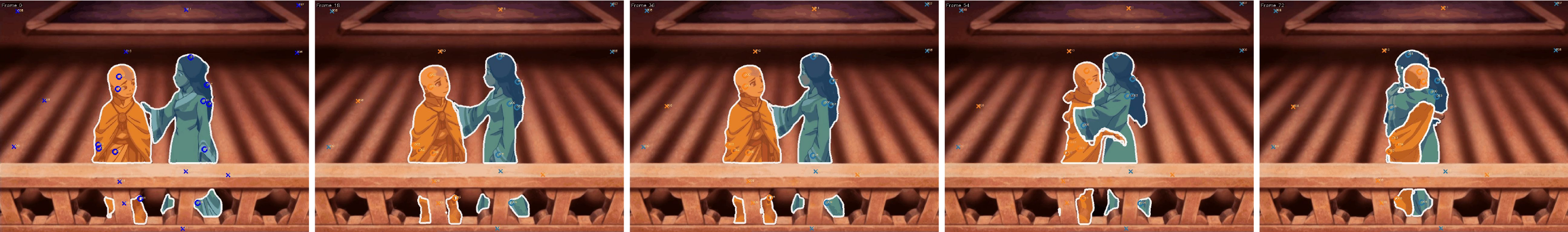}
    \vskip0.3mm
    \includegraphics[width=\linewidth]{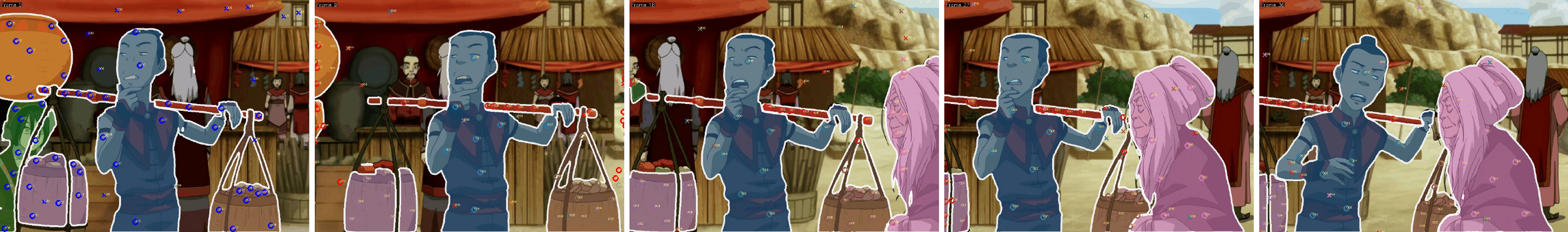}
    \vspace{-0.28in}
    \caption{{Successful segmentation using SAM-PT on short clips from ``Avatar: The Last Airbender''.} Although our method has never seen data from Avatar, an anime-influenced animated television series, it segments and tracks various objects in short clips.}
    \label{fig:avatar-good-cases}
    \vspace{-0.2in}
\end{figure*}

\vspace{0.5mm}
\noindent\textbf{Qualitative Analysis.}\quad
Our visualizations in~\cref{davis-good-cases-samp} and \cref{davis-good-cases-samp-reinit} demonstrate successful segmentation on the DAVIS $2017$ dataset and underscore our method's ability to perform zero-shot video segmentation on unseen content, such as clips from the anime-influenced series ``Avatar: The Last Airbender'' in~\cref{fig:avatar-good-cases}. These examples highlight the versatility and adaptability of SAM-PT.

\noindent \textbf{Limitations and Challenges.}\quad
Our method excels in zero-shot video object segmentation but faces challenges with point tracker reliability in complex scenarios, such as occlusions and fast-moving objects, as shown in \cref{davis-failure-cases-samp}.
While point reinitialization and negative point strategies offer some improvement, interactive use significantly bridges the performance gap with trained methods. This interactive potential is successfully demonstrated in \cref{sec:exp-interaction-results}, where user intervention enhances segmentation accuracy.

\subsection{Video Instance Segmentation}
\label{sec:exp-results-vis}
Given the same mask proposals, SAM-PT outperforms TAM~\cite{yang2023track} significantly, as shown in \cref{tab:uvo-v1-val}, even though SAM-PT was not trained on any video segmentation data. TAM is a concurrent approach combining SAM and XMem~\cite{cheng2022xmem}, where XMem was pre-trained on BL30K~\cite{cheng2021modular} and trained on DAVIS and YouTube-VOS, but not on UVO. On the other hand, SAM-PT combines SAM with the PIPS or CoTracker point tracking method, both of which have not been trained on video segmentation tasks.

\subsection{Interactive Point-Based Video Segmentation}
\label{sec:exp-interaction-results}
\begin{figure}[h]
\centering
\includegraphics[width=0.80\linewidth]{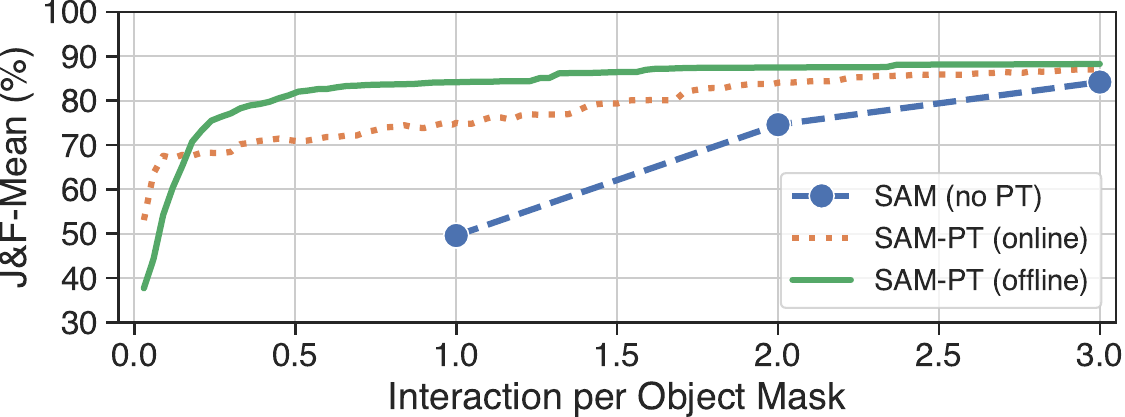}
\vspace{-0.1in}
\caption{Interactive segmentation performance on the DAVIS 2017~\cite{ponttuset2018davis} validation set. SAM-PT's offline strategy notably outperforms the baseline SAM, demonstrating efficient video annotation with minimal user intervention.}
\label{fig:interactive}
\vspace{-0.15in}
\end{figure}

Building upon our method's strengths observed in standard benchmarks, this %
study evaluates the responsiveness of SAM-PT to human input, aiming to understand its practical utility for interactive video annotation tasks. We benchmarked SAM-PT's interactive performance against a baseline SAM approach that does not utilize point tracking. The results, visualized in \cref{fig:interactive}, show that SAM-PT significantly outperforms the baseline, particularly when employing the offline checkpoint strategy. Although this offline method initially starts at a lower performance due to setup costs, it quickly exceeds the online method's performance by benefiting from a cumulative optimization approach.

These results suggest that SAM-PT substantially reduces the effort required for high-quality video annotation, bringing its performance closer to fully-supervised methods and highlighting its practical utility.

\input{tables/bdd100k-vos-valid}

\input{tables/uvo-dense-video-v1.0-validation}

\section{Conclusion}
\label{sec:conclusion}
SAM-PT introduces a point-centric approach for interactive video segmentation by combining the generalization capabilities of the Segment Anything Model (SAM) and long-term point tracking. Our work fills in the gap that the point-centric approach is scantly explored in the literature. In our experiments, SAM-PT achieves strong performance across video segmentation tasks including semi-supervised, open-world, and fully interactive video segmentation. While our method has limitations such as difficulty handling occlusions, small objects, motion blur, and inconsistencies in mask predictions, it contributes a new perspective to video object segmentation research.

{
    \small
    \bibliographystyle{ieeenat_fullname}
    \bibliography{bibtex/literature}
}

\appendix
\maketitlesupplementary
In this supplementary, we first report experimental results on additional datasets and subsets (\cref{supp-mose-2023,supp-davis-2016,supp-davis-2017-testdev}). Then we extend and detail our ablation and report more qualitative results (\cref{supp-sam-backbones,supp-sam-variants,supp-more-ablation-experiments,supp-reinit-details,supp-qual-results}). Lastly, we detail on our evaluation protocols (\cref{supp-vos-eval-details,supp-vis-eval-details,supp-bdd100k-creation,supp-interactive-details}). %

\section{MOSE 2023}
\label{supp-mose-2023}
MOSE $2023$~\cite{ding2023mose} is a recently introduced dataset that focuses on multi-object segmentation and tracking in complex scenes, replete with challenges such as occlusions, transient visibility of objects, extensive occlusion, etc. Our results on the validation subset suggest that SAM-PT achieves performance competitive with SegGPT~\cite{wang2023seggpt}, as shown in \cref{table:vos-mose-2023-val}.
\input{tables/mose-2023-valid}

\section{DAVIS 2016}
\label{supp-davis-2016}
DAVIS $2016$ offers a single-object VOS benchmark across $20$ diverse sequences. We report results on the DAVIS $2016$ validation subset in \cref{table:vos-davis-2016-val}, in which our method achieves $84.3$ points, surpassing SegGPT's $82.3$ points.
\input{tables/davis-2016-valid}

\section{DAVIS 2017 Test-dev Subset}
\label{supp-davis-2017-testdev}
DAVIS $2017$ is a multi-object extension of its $2016$ version. The video scenarios within this dataset are small but diverse. In addition to the results on the validation subset in the main manuscript, we report the performance of SAM-PT on the DAVIS $2017$ test-dev subset in \cref{table:vos-davis-2017-testdev}.
\input{tables/davis-2017-testdev}

\section{Different SAM Backbones}
\label{supp-sam-backbones}
The SAM model's backbone plays an important role in determining its performance and inference speed. In this experiment, we evaluated SAM with different ViT backbones: ViT-Huge (used throughout the work), ViT-Large, and ViT-Base. The results, as measured on the validation subset of DAVIS $2017$ are shown in \cref{tab:sam-backbones}. Replacing ViT-Huge with ViT-Large results in only a non-significant loss in performance, SAM-PT's overall performance nevertheless remains non-real-time. Note that the ViT-Huge number in this experiment is slightly different from the one presented in the main manuscript due to the use of different seed values.
\input{tables/ablation-sam-backbones}

\section{Different SAM Variants}
\label{supp-sam-variants}
To cater to scenarios where inference speed is crucial, we explored lightweight variants of SAM: Light HQ-SAM~\cite{ke2023segment} and MobileSAM~\cite{zhang2023faster}. The performance and speed trade-offs for these variants are summarized in table \cref{tab:sam-lightweight}. Using the HQ-SAM~\cite{ke2023segment} variant of SAM results in the highest performance of $77.64$ points, whereas MobileSAM has the highest inference speed of $5.5$ FPS. Using the lightweight variants doesn't achieve real-time performance as the bottleneck of the pipeline moves to the point tracker.
\input{tables/ablation-sam-lightweight-variants}

\section{Detailed Ablation Results}
\label{supp-more-ablation-experiments}
We report detailed experimental results of our ablation studies for configurations using PIPS~\cite{harley2022particle} as the point tracker in \cref{table:ablation-pips-detailed} and using CoTracker~\cite{karaev2023cotracker} in \cref{table:ablation-cotracker}.
\input{tables/ablation-cotracker}
\input{tables/ablation-pips-detailed}

\section{Point Tracking Reinitialization}
\label{supp-reinit-details}
In our method, we introduce an optional reinitialization strategy. Here, the point tracker begins anew every $h$ frames, where $h$ represents a pre-set tracking \textit{horizon} (e.g., $8$ frames), or is dynamically determined based on SAM's mask predictions for each timestep within the horizon (e.g., using most-similar-mask-area heuristics). Upon reaching this horizon, the query points given to the tracker are reinitialized according to the mask prediction SAM outputted at the horizon frame. While this method may increase the computational load, it shows performance improvement when using the PIPS~\cite{harley2022particle} point tracker in demanding video sequences, such as those in the MOSE dataset. However, our studies also suggested that the proposed reinitialization strategies hurt performance when using CoTracker as the point tracker. We primarily designed the reinitialization variants to address some of the failure cases of PIPS such as the common case of points being wrongly predicted to be off the target object, being predicted on the background instead, but this does not work as well for other point trackers such as CoTracker that are more robust to such failures.

We explored four reinitialization strategies, each varying in how they compute the value of $h$:
\begin{enumerate}[label=(\textbf{\Alph*})]
\item \textbf{Reinit-on-Horizon-and-Sync-Masks}: This straightforward variant reinitializes points after a fixed number of frames (e.g., every 8 frames). However, it may stumble if the mask is absent at the reinitialization timestep.
\item \textbf{Reinit-at-Median-of-Area-Diff}: In this variant, the tracker outputs trajectory points for each frame within the horizon, and SAM predicts masks based on these trajectories. Reinitialization happens at the frame within the horizon that has the mean mask area among the non-empty masks predicted by SAM.
\item \textbf{Reinit-on-Similar-Mask-Area}: This method triggers reinitialization when the mask area is similar to the initial mask area.
\item \textbf{Reinit-on-Similar-Mask-Area-and-Sync-Masks}: This variant reinitializes when the mask area for all masks in the batch is similar to the initial mask areas, synchronizing the masks to be tracked from the same timestep. This synchronization allows for the use of negative points from other masks when querying SAM.
\end{enumerate}

From our ablation investigations, we found the \textbf{(A) Reinit-on-Horizon-and-Sync-Masks} strategy to be effective with PIPS as the point tracker and \textbf{(B) Reinit-at-Median-of-Area-Diff} with CoTracker. Besides the point tracker used, the choice of reinitialization method may depend on the specific validation subset and the degree of hyperparameter tuning involved. Note that we always use reinitialization along with negative points.

\section{Additional Qualitative Results}
\label{supp-qual-results}

\begin{figure*}[p]
    \centering
        \begin{subfigure}{\linewidth}
        \centering
        \includegraphics[width=\linewidth]{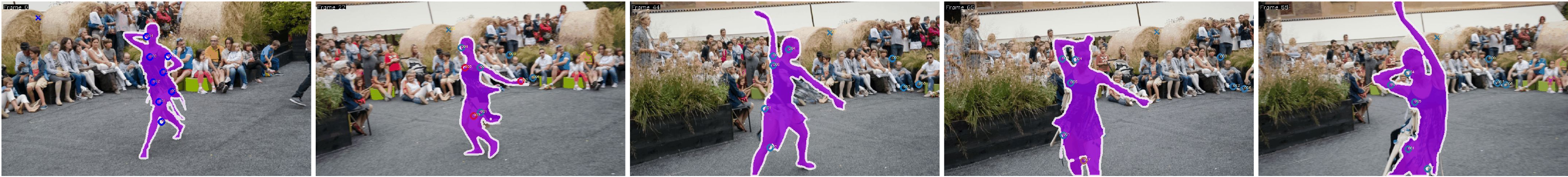}
        \includegraphics[width=\linewidth]{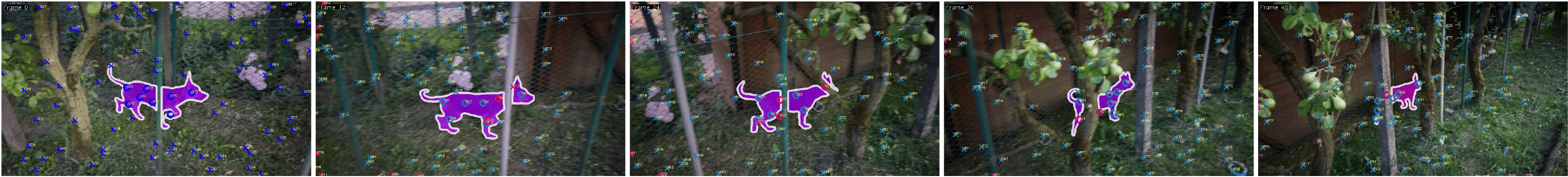}
        \includegraphics[width=\linewidth]{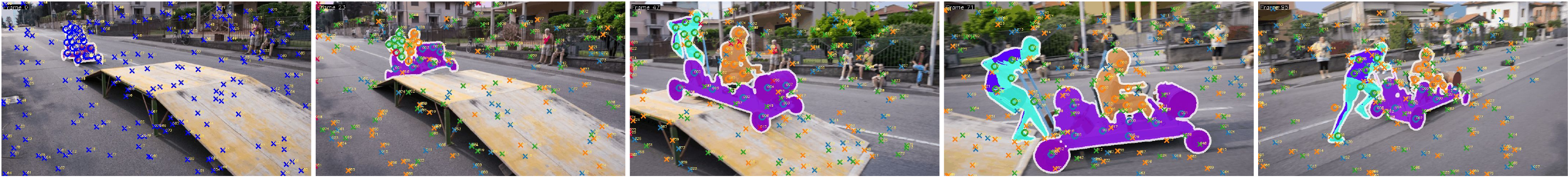}
        \caption{Successful cases for SAM-PT.}
    \end{subfigure}
    \centering
        \begin{subfigure}{\linewidth}
        \centering
        \vspace{0.12cm}
        \includegraphics[width=\linewidth]{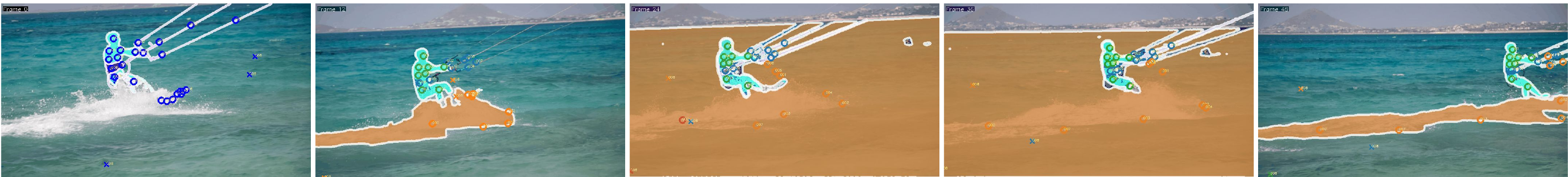}
        \includegraphics[width=\linewidth]{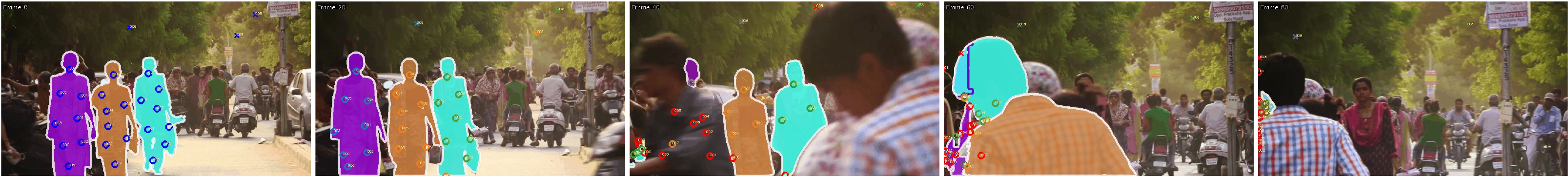}
        \caption{Failure cases for SAM-PT.}
    \end{subfigure}
    \vspace{-0.28in}
    \caption{Additional visualization of SAM-PT on videos from the DAVIS 2017~\cite{ponttuset2018davis} validation subset, including (a) successful cases and (b) failure cases. Circles denote positive points and crosses denote negative points. Red symbols indicate occlusion prediction.
    }
    \label{fig:additional-davis-qualitative-results}
\end{figure*}

\begin{figure*}[p]
    \centering
    \includegraphics[width=\linewidth]{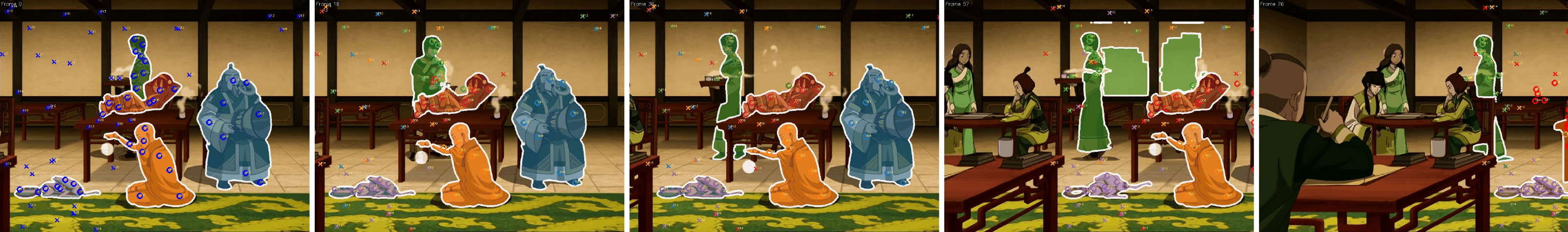}
    \vskip0.3mm
    \includegraphics[width=\linewidth]{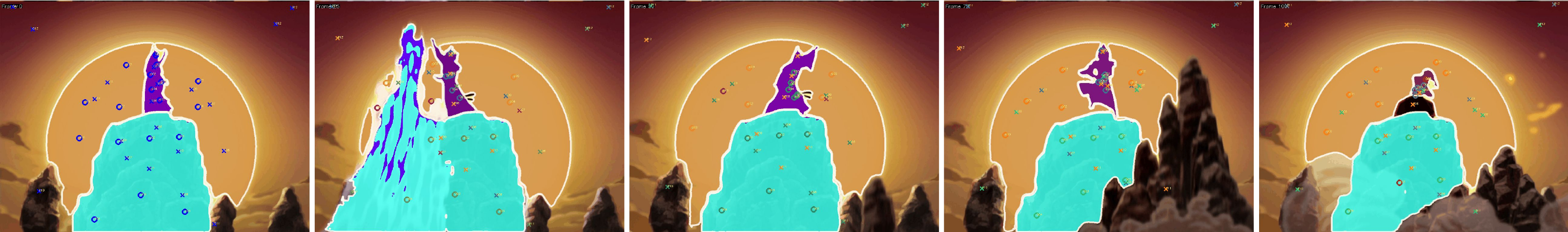}
    \vskip0.3mm
    \includegraphics[width=\linewidth]{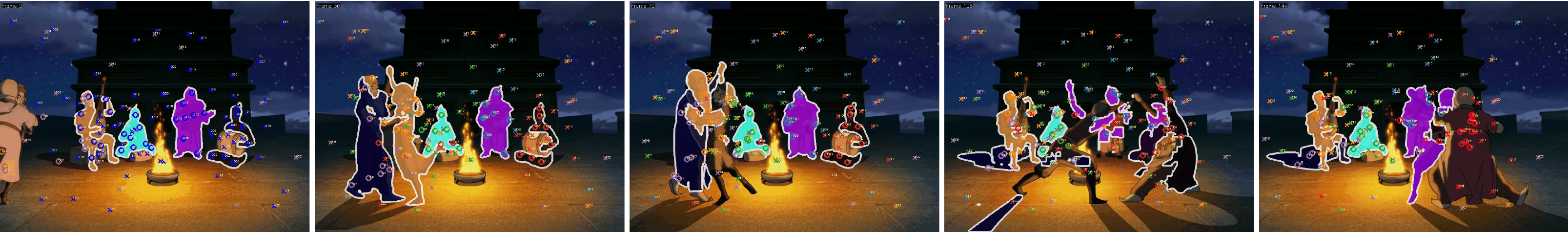}
    \vspace{-0.28in}
    \caption{{Challenging scenarios for SAM-PT on short clips from ``Avatar: The Last Airbender''.} These cases illustrate instances where our model struggles when faced with point tracking failures that are the result of incorrectly predicting the point at a similar-looking segment or when faced with object occlusions and disappearing objects.}
    \label{fig:avatar-failure-cases}
    \vspace{-0.2in}
\end{figure*}

We show additional visualizations on DAVIS videos in \cref{fig:additional-davis-qualitative-results}. We show failure cases on clips from the anime-influenced series ``Avatar: The Last Airbender'' in \cref{fig:avatar-failure-cases}.

\section{VOS Evaluation Details}
\label{supp-vos-eval-details}
When evaluating on VOS, we use the provided ground truth mask for the first frame to sample the query points required by our method. Then, we give only the sampled points as input to our method, not the mask. For all datasets, we use the full-resolution data and resize it to the longest side of $1024$ to match SAM's input resolution.

\section{VIS Evaluation Details}
\label{supp-vis-eval-details}
For evaluating our method on the VIS task, we leverage SAM's automatic mask generation capacity to generate up to 100 mask proposals for the initial frame. We use the same initial masks to assess TAM~\cite{yang2023track} for a fair comparison. Our current approach does not generate new proposals in subsequent frames, thus it cannot detect new objects appearing after the first frame, unlike fully-fledged VIS methods. However, this setup allows for a straightforward comparison of zero-shot capabilities in mask propagation from the initial frame.

\section{BDD100K VOS Dataset Creation}
\label{supp-bdd100k-creation}

BDD100K is a large open driving video dataset with 100K videos and 10 tasks to evaluate the progress of image recognition algorithms in autonomous driving. It includes a variety of geographic, environmental, and weather conditions. We convert its annotations from the Multi-Object Tracking and Segmentation (MOTS) section into semi-supervised VOS annotations. This section has $154$, $32$, and $37$ videos for train, validation, and test sets, totaling $25$K instances and $480$K masks.

To convert the annotations, we take the ground truth mask of the first appearance of each object in the videos. This mask will be given as input to semi-supervised VOS methods which then need to predict the masks for the remaining video frames. In converting the validation subset of $32$ videos and $4566$ object tracks, we create $61$ semi-supervised VOS datapoints of up to $100$ objects per video. We limit the number of objects per video for implementation simplicity since most VOS methods expect a small number of objects per video. For example, in the DAVIS validation subset, the maximum number of objects per video is 5. During the conversion, we additionally remove the instances marked as ``ignored'' or ``crowd'' in the MOTS annotations.

\section{Interactive Point-Based Video Segmentation Details}
\label{supp-interactive-details}
Interactive point-based video segmentation aims to refine segmentation masks with minimal user input while optimizing the Intersection over Union (IoU) with the ground truth. To evaluate the responsiveness of SAM-PT to simulated human input, we benchmark against a SAM-only baseline (\cref{alg:sam-only}) and compare it with both an online (\cref{alg:online-sampt}) and offline (\cref{alg:offline-sampt}) interactive SAM-PT method:
\begin{enumerate}
    \item \textbf{Non-tracking method (SAM only):} As a baseline, this method mimics user interactions by selecting points on each frame without any point tracking, effectively emulating the process of manual, frame-by-frame annotation using the standalone SAM model.
    \item \textbf{Online method:} This approach models a user going through the video sequentially a single time, making corrective per-frame interactions to achieve an IoU of at least 95\% with the ground truth mask for each frame. These corrections include the addition or removal of points and are propagated to subsequent frames via point tracking. The user may opt to skip frames that already appear to be well-annotated or that cannot be annotated sufficiently well within the available interaction budget.
    \item \textbf{Offline method (Checkpoint Strategy):} This method takes a multi-pass approach, incrementally aiming for higher IoU thresholds with each pass through the video. Checkpoints are saved after each pass, and the latest checkpoint given the interaction budget is returned, allowing for a more global optimization approach.
\end{enumerate}
Interactions are defined as the act of adding or removing a point and are executed as described in \cref{alg:perform-int}. To maintain simplicity in our evaluation, skipping frames while progressing through a video is not counted as an interaction.
\vspace{0.1in}

\begin{algorithm}
\caption{Non-tracking (SAM only) Method.}
\label{alg:sam-only}
\begin{algorithmic}[1]
\STATE \textbf{Input:} rgbs, gt\_masks, int\_per\_frame
\STATE \textbf{Output:} pred\_masks
\STATE point\_memory $\leftarrow$ emptyPointMemory()
\FOR{i in frames}
    \STATE pmf $\leftarrow$ emptyFramePointMemory()
    \FOR{j in int\_per\_frame}
        \STATE m $\leftarrow$ predMask(rgbs[i], pmf)
        \STATE performInt(i, m, gt\_masks[i], pmf)
    \ENDFOR
    \STATE addToPointMemory(point\_memory, i, pmf)
\ENDFOR
\RETURN predAllMasks(rgbs, point\_memory)
\end{algorithmic}
\end{algorithm}

\begin{algorithm}
\caption{Online Method.}
\label{alg:online-sampt}
\begin{algorithmic}[1]
\STATE \textbf{Input:} rgbs, gt\_masks
\STATE \textbf{Output:} pred\_masks
\STATE max\_int $\leftarrow 300$
\STATE max\_int\_per\_frame $\leftarrow 3$
\STATE threshold $\leftarrow 0.95$
\STATE point\_memory $\leftarrow$ selectFirstPoint(gt\_masks[0])
\STATE max\_int $\leftarrow$ max\_int - 1
\FOR{i in frames}
    \STATE m $\leftarrow$ predMask(rgbs[i], point\_memory)
    \STATE IoU $\leftarrow$ calculateIoU(m, gt\_masks[i])
    \IF{IoU $\geq$ threshold}
        \STATE continue
    \ENDIF
    \STATE performInt(i, m, gt\_masks[i], point\_memory)
    \STATE max\_int $\leftarrow$ max\_int - 1
    \IF{max\_int $\leq 0$}
        \STATE break
    \ENDIF
\ENDFOR
\RETURN predAllMasks(rgbs, point\_memory)
\end{algorithmic}
\end{algorithm}

\begin{algorithm}
\caption{Offline Method (Checkpoint Strategy).}
\label{alg:offline-sampt}
\begin{algorithmic}[1]
\STATE \textbf{Input:} rgbs, gt\_masks
\STATE \textbf{Output:} pred\_masks
\STATE max\_int $\leftarrow 300$
\STATE max\_int\_per\_frame $\leftarrow 3$
\STATE int\_iou\_thresholds $\leftarrow [0.10, 0.20, \ldots, 0.95]$
\STATE point\_memory $\leftarrow$ selectFirstPoint(gt\_masks[0])
\STATE max\_int $\leftarrow$ max\_int - 1
\STATE best\_ckpt $\leftarrow$ predAllMasks(rgbs, point\_memory)
\FOR{threshold in int\_iou\_thresholds}
    \FOR{i in frames}
        \FOR{j in max\_int\_per\_frame}
            \STATE m $\leftarrow$ predMask(rgbs[i], point\_memory)
            \STATE IoU $\leftarrow$ calculateIoU(m, gt\_masks[i])
            \IF{IoU $\geq$ threshold}
                \STATE break
            \ENDIF
            \STATE performInt(i, m, gt\_masks[i], point\_memory)
            \STATE max\_int $\leftarrow$ max\_int - 1
        \ENDFOR
    \ENDFOR
    \IF{max\_int $\geq 0$}
        \STATE best\_ckpt $\leftarrow$ predAllMasks(rgbs, point\_memory)
    \ELSE
        \STATE break
    \ENDIF
\ENDFOR
\RETURN best\_ckpt
\end{algorithmic}
\end{algorithm}

\begin{algorithm}
\caption{Perform Interaction.}
\label{alg:perform-int}
\begin{algorithmic}[1]
\STATE \textbf{Input:} frame\_idx, m, gt\_m, point\_memory
\STATE i $\gets$ frame\_idx
\STATE tp\_mask $\leftarrow$ m $\land$ gt\_m
\STATE tn\_mask $\leftarrow$ $\lnot$m $\land$ $\lnot$gt\_m
\STATE fp\_mask $\leftarrow$ m $\land$ $\lnot$gt\_m
\STATE fn\_mask $\leftarrow$ $\lnot$m $\land$ gt\_m
\STATE pos\_points $\leftarrow$ getPosPoints(point\_memory, i)
\STATE neg\_points $\leftarrow$ getNegPoints(point\_memory, i)
\IF{any neg\_points in fn\_mask}
    \STATE p $\leftarrow$ firstIncorrectNegPoint(fn\_mask, neg\_points)
    \STATE removePointAndItsFuture(point\_memory, i, p)
\ELSIF{any pos\_points in fp\_mask}
    \STATE p $\leftarrow$ firstIncorrectPosPoint(fp\_mask, pos\_points)
    \STATE removePointAndItsFuture(point\_memory, i, p)
\ELSE
    \IF{sum(fn\_mask) $>$ sum(fp\_mask)}
        \STATE [x, y] $\leftarrow$ extractPoint(fn\_mask)
        \STATE lbl $\leftarrow$ positive
    \ELSE
        \STATE [x, y] $\leftarrow$ extractPoint(fp\_mask)
        \STATE lbl $\leftarrow$ negative
    \ENDIF
    \STATE addToFrameAndFuture(point\_memory, i, lbl, x, y)
\ENDIF
\end{algorithmic}
\end{algorithm}

\end{document}

%% file: preamble.tex
\usepackage{multirow}
\usepackage[nolist,nohyperlinks]{acronym}
\usepackage{dblfloatfix}
\usepackage{boldline}
\usepackage{bm}
\usepackage{pifont}
\newcommand{\gr}[1]{{\textcolor{gray}{#1}}}
\newlength\savewidth

\usepackage{makecell}

\usepackage{bbding}
\usepackage{color}
\usepackage{pifont}
\usepackage{xcolor}
\usepackage{colortbl}

\usepackage{enumitem}
\setlist{nosep} %

\usepackage{caption}
\usepackage{subcaption}
\usepackage[skip=0.5ex]{subcaption}

\usepackage{xspace}

\usepackage{arydshln}
\newcommand{\dashrule}[1][black]{%
  \color{#1}\rule[\dimexpr.5ex-.2pt]{4pt}{.4pt}\xleaders\hbox{\rule{4pt}{0pt}\rule[\dimexpr.5ex-.2pt]{4pt}{.4pt}}\hfill\kern0pt%
}
\newcommand{\rulecolor}[1]{%
  \def\CT@arc@{\color{#1}}%
}

\usepackage{cuted}
\usepackage{capt-of}

\usepackage{algorithm}
\usepackage{algorithmic}

%% file: tables/ablation-point-trackers.tex
\begin{table}
\centering
\caption{
We report the mean performance and standard deviation across eight runs on the validation subset of DAVIS $2017$ to study the impact of different point trackers.
}
\vspace{-0.07in}
\label{table:ablation-point-trackers}
\resizebox{0.7\columnwidth}{!}{
\begin{tabular}{lcccc}
\toprule
\multirow{2}{*}{\textbf{Point Tracker}}& \multicolumn{3}{c}{\textbf{DAVIS 2017 Validation}~\cite{ponttuset2018davis}} \\
\cmidrule{2-5}
& \textbf{$\mathcal{J\&F}$ ↑} & \textbf{$\mathcal{J}$ ↑} & \textbf{$\mathcal{F}$ ↑} \\
\toprule
SuperGlue~\cite{sarlin2020superglue}  & $28.4$\gr{\scriptsize{$\pm 3.1$}} & $24.7$\gr{\scriptsize{$\pm 2.4$}} & $32.0$\gr{\scriptsize{$\pm 3.8$}} \\
TapNet~\cite{doersch2023tapvid}       & $60.9$\gr{\scriptsize{$\pm 0.2$}} & $58.2$\gr{\scriptsize{$\pm 0.3$}} & $63.5$\gr{\scriptsize{$\pm 0.2$}} \\
RAFT~\cite{teed2020raft}              & $63.0$\gr{\scriptsize{$\pm 0.6$}} & $60.7$\gr{\scriptsize{$\pm 0.6$}} & $65.4$\gr{\scriptsize{$\pm 0.5$}} \\
PIPS++~\cite{zheng2023point}          & $73.2$\gr{\scriptsize{$\pm 0.5$}} & $69.9$\gr{\scriptsize{$\pm 0.5$}} & $76.6$\gr{\scriptsize{$\pm 0.5$}} \\
PIPS~\cite{harley2022particle}        & $76.3$\gr{\scriptsize{$\pm 0.6$}} & $73.6$\gr{\scriptsize{$\pm 0.6$}} & $78.9$\gr{\scriptsize{$\pm 0.6$}} \\
TAPIR~\cite{doersch2023tapir}         & $76.7$\gr{\scriptsize{$\pm 0.3$}} & $73.8$\gr{\scriptsize{$\pm 0.4$}} & $79.7$\gr{\scriptsize{$\pm 0.3$}} \\
\rowcolor{lightgray!16}
CoTracker~\cite{karaev2023cotracker}  & $\mathbf{77.6}$\gr{\scriptsize{$\pm 0.7$}} & $\mathbf{74.8}$\gr{\scriptsize{$\pm 0.7$}} & $\mathbf{80.4}$\gr{\scriptsize{$\pm 0.7$}} \\
\bottomrule
\end{tabular}
}
\vspace{-0.2in}
\end{table}

%% file: tables/ablation-pips.tex
\begin{table}
\centering
\caption{
    Ablation study on the validation subset of DAVIS $2017$ on the impact of different SAM-PT configurations using PIPS~\cite{harley2022particle} as the point tracker.
    PSM: point selection method.
    PP: positive points per mask.
    NP: negative points per mask.
    IRI: iterative refinement iterations.
    PS: point similarity filtering.
    RV: reinitalization variant.
}
\vspace{-0.07in}
\label{table:ablation-pips}
\resizebox{0.84\columnwidth}{!}{
\begin{tabular}{cccccccr}
\toprule
\multicolumn{6}{c}{\textbf{SAM-PT Configuration (using PIPS)}} & \multicolumn{1}{c}{\textbf{DAVIS}~\cite{ponttuset2018davis}} \\
\cmidrule{7-7}
{PSM} & {PP} & {NP} & {IRI} & {PS} &{RV} & \textbf{$\mathcal{J\&F}$ ↑} & Gain        \\ \toprule
    Random & $ 1$  & $0$  & $0$  & \ding{55} & \ding{55} & $37.1$\gr{\scriptsize{$\pm 21.7$}} &         \\
    Random & $ 8$  & $0$  & $0$  & \ding{55} & \ding{55} & $70.5$\gr{\scriptsize{$\pm  1.4$}} & $+33.4$ \\ \midrule
 K-Medoids & $ 8$  & $0$  & $0$  & \ding{55} & \ding{55} & $72.3$\gr{\scriptsize{$\pm  1.2$}} &  $+1.8$ \\
Shi-Tomasi & $ 8$  & $0$  & $0$  & \ding{55} & \ding{55} & $72.0$\gr{\scriptsize{$\pm  0.3$}} &         \\
     Mixed & $ 8$  & $0$  & $0$  & \ding{55} & \ding{55} & $70.6$\gr{\scriptsize{$\pm  0.8$}} &         \\ \midrule
 K-Medoids & $ 8$  & $1$  & $0$  & \ding{55} & \ding{55} & $74.1$\gr{\scriptsize{$\pm  0.7$}} &  $+1.8$ \\ \midrule
 K-Medoids & $ 8$  & $1$  & $12$ & \ding{55} & \ding{55} & $76.3$\gr{\scriptsize{$\pm  0.6$}} &  $+2.2$ \\ \midrule
 K-Medoids & $ 8$  & $1$  & $12$ & \ding{51} & \ding{55} & $72.7$\gr{\scriptsize{$\pm  2.0$}} &    none \\ \midrule
 K-Medoids & $ 8$  & $72$ & $12$ & \ding{55} &         A & $76.8$\gr{\scriptsize{$\pm  0.7$}} &  $+0.5$ \\
 K-Medoids & $ 8$  & $ 1$ & $12$ & \ding{55} &         B & $76.1$\gr{\scriptsize{$\pm  0.4$}} &         \\
 K-Medoids & $ 8$  & $ 1$ & $ 0$ & \ding{55} &         C & $75.5$\gr{\scriptsize{$\pm  0.7$}} &         \\
 K-Medoids & $ 8$  & $ 1$ & $12$ & \ding{55} &         D & $76.4$\gr{\scriptsize{$\pm  0.3$}} &         \\
\bottomrule
\end{tabular}
}
\vspace{-0.15in}
\end{table}

%% file: tables/davis-2017-valid.tex
\begin{table}
\centering
\caption{Quantitative results on the DAVIS $2017$~\cite{ponttuset2018davis} validation set for semi-supervised VOS. Performance is reported for different methods, including our SAM-PT and HQ-SAM-PT, with and without the reinitialization strategy (Reinit) and using different point trackers. Our method outperforms other zero-shot methods.}
\vspace{-0.1in}
\label{table:vos-davis-2017-val}
\resizebox{\columnwidth}{!}{
\begin{tabular}{llccccc}
\toprule
& \multirow{2}{*}{\textbf{Method}} & \multirow{2}{*}{\textbf{Tracker}} & \multirow{2}{*}{\textbf{Reinit}} & \multicolumn{3}{c}{\textbf{DAVIS $\mathbf{2017}$ Validation}~\cite{ponttuset2018davis}}
\\
\cmidrule{5-7}
& & & & \textbf{$\mathcal{J\&F}$ ↑} & \textbf{$\mathcal{J}$ ↑} & \textbf{$\mathcal{F}$ ↑}
\\
\toprule
\multicolumn{7}{l}{(a) trained on video segmentation data} \\
\midrule
& \gr{MiVOS}~\cite{cheng2021mivos}             & \gr{-} & \gr{-} & \gr{$84.5$} & \gr{$81.7$} & \gr{$87.4$} \\
& \gr{DeAOT~\cite{yang2022decoupling}}         & \gr{-} & \gr{-} & \gr{$86.2$} & \gr{$83.1$} & \gr{$89.2$} \\
& \gr{DEVA~\cite{cheng2023tracking}}           & \gr{-} & \gr{-} & \gr{$87.6$} & \gr{$84.2$} & \gr{$91.0$} \\
& \gr{XMem~\cite{cheng2022xmem}}               & \gr{-} & \gr{-} & \gr{$87.7$} & \gr{$84.0$} & \gr{$91.4$} \\
\midrule
\multicolumn{5}{l}{(b) not trained on video segmentation data (\textit{\textbf{zero-shot}})} \\
\midrule
& Painter~\cite{wang2023images}        & - & - & $34.6$ & $28.5$ & $40.8$ \\
& DINO~\cite{caron2021emerging}        & - & - & $71.4$ & $67.9$ & $74.9$ \\
& SegGPT~\cite{wang2023seggpt}         & - & - & ${75.6}$ & ${72.5}$ & ${78.6}$ \\ \midrule
& \multirow{4}{*}{\makecell[l]{{SAM-PT}\\{(ours)}}}
  & PIPS~\cite{harley2022particle}       & \ding{55} & $76.3$\gr{\scriptsize{$\pm 0.6$}}  & $73.6$\gr{\scriptsize{$\pm 0.6$}}  & ${78.9}$\gr{\scriptsize{$\pm 0.6$}} \\
& & PIPS~\cite{harley2022particle}       & \ding{51} & ${76.6}$\gr{\scriptsize{$\pm 0.7$}}  & ${74.4}$\gr{\scriptsize{$\pm 0.8$}}  & ${78.9}$\gr{\scriptsize{$\pm 0.6$}} \\
& & CoTracker~\cite{karaev2023cotracker} & \ding{55} & ${77.6}$\gr{\scriptsize{$\pm 0.7$}} & ${74.8}$\gr{\scriptsize{$\pm 0.7$}} & ${80.4}$\gr{\scriptsize{$\pm 0.7$}} \\
& & CoTracker~\cite{karaev2023cotracker} & \ding{51} & $77.4$\gr{\scriptsize{$\pm 1.0$}} & $74.5$\gr{\scriptsize{$\pm 1.0$}} & $80.3$\gr{\scriptsize{$\pm 1.1$}} \\ \midrule
& \multirow{4}{*}{\makecell[l]{{HQ-SAM-PT}\\{(ours)}}}
  & PIPS~\cite{harley2022particle}       & \ding{55} &  $77.2$\gr{\scriptsize{$\pm 0.5$}}  & $74.7$\gr{\scriptsize{$\pm 0.5$}}  & $79.8$\gr{\scriptsize{$\pm 0.4$}} \\
& & PIPS~\cite{harley2022particle}       & \ding{51} &  $77.0$\gr{\scriptsize{$\pm 0.7$}}  & $74.8$\gr{\scriptsize{$\pm 0.8$}}  & $79.2$\gr{\scriptsize{$\pm 0.6$}} \\
& & CoTracker~\cite{karaev2023cotracker} & \ding{55} &  $\textbf{79.4}$\gr{\scriptsize{$\pm 0.6$}} & $\textbf{76.5}$\gr{\scriptsize{$\pm 0.6$}} & $\textbf{82.3}$\gr{\scriptsize{$\pm 0.5$}} \\
& & CoTracker~\cite{karaev2023cotracker} & \ding{51} &  $77.7$\gr{\scriptsize{$\pm 0.8$}} & $74.6$\gr{\scriptsize{$\pm 0.9$}} & $80.8$\gr{\scriptsize{$\pm 0.7$}} \\
\bottomrule
\end{tabular}
}
\end{table}

%% file: tables/ytvos-2018-valid.tex
\begin{table}
\centering
\caption{Quantitative results in semi-supervised VOS on the validation subset of YouTube-VOS $2018$. Metrics are reported separately for ``seen'' and ``unseen'' classes, with $\mathcal{G}$ being the overall average score over the metrics.
}
\label{table:vos-ytvos-2018-val}
\vspace{-0.15in}
\resizebox{\columnwidth}{!}{
\begin{tabular}{llccccccc}
\toprule
& \multirow{2}{*}{\textbf{Method}} & \multirow{2}{*}{\textbf{Tracker}} & \multirow{2}{*}{\textbf{Reinit}} & \multicolumn{5}{c}{\textbf{YouTube-VOS $\mathbf{2018}$ Validation}~\cite{xu2018youtubevos}} \\
\cmidrule{5-9}
& & &  & $\mathcal{G}$ & $\mathcal{J}_s$ & $\mathcal{F}_s$ & $\mathcal{J}_u$ & $\mathcal{F}_u$
\\
\toprule
\multicolumn{7}{l}{(a) trained on video segmentation data} \\
\midrule
& \gr{XMem~\cite{cheng2022xmem}}               & \gr{-} & \gr{-} & \gr{$86.1$} & \gr{$85.1$} & \gr{$89.8$} & \gr{$80.3$} & \gr{$89.2$} \\
& \gr{DeAOT~\cite{yang2022decoupling}} & \gr{-} & \gr{-} & \gr{$86.2$} & \gr{$85.6$} & \gr{$90.6$} & \gr{$80.0$} & \gr{$88.4$} \\

\midrule
\multicolumn{7}{l}{(b) not trained on video segmentation data (\textit{\textbf{zero-shot}})} \\
\midrule
& Painter~\cite{wang2023images} & - & - & $24.1$ & $27.6$ & $35.8$ & $14.3$ & $18.7$ \\
& SegGPT~\cite{wang2023seggpt}  & - & - & ${74.7}$ & ${75.1}$ & $\mathbf{80.2}$ & ${67.4}$ & ${75.9}$ \\
\midrule
& \multirow{4}{*}{\makecell[l]{SAM-PT\\(ours)}}
  & PIPS~\cite{harley2022particle}       & \ding{55} & $67.0 \textcolor{gray}{\scriptsize{\pm 0.3}}$ & $68.6 \textcolor{gray}{\scriptsize{\pm 0.2}}$ & $71.2 \textcolor{gray}{\scriptsize{\pm 0.1}}$ & $61.0 \textcolor{gray}{\scriptsize{\pm 0.5}}$ & $67.4 \textcolor{gray}{\scriptsize{\pm 0.4}}$ \\
& & PIPS~\cite{harley2022particle}       & \ding{51} & ${67.5} \textcolor{gray}{\scriptsize{\pm 0.2}}$ & ${69.0} \textcolor{gray}{\scriptsize{\pm 0.4}}$ & ${69.9} \textcolor{gray}{\scriptsize{\pm 0.3}}$ & ${63.2} \textcolor{gray}{\scriptsize{\pm 0.4}}$ & ${67.8} \textcolor{gray}{\scriptsize{\pm 0.5}}$ \\
& & CoTracker~\cite{karaev2023cotracker} & \ding{55} & $74.0 \textcolor{gray}{\scriptsize{\pm 0.3}}$ & $73.3 \textcolor{gray}{\scriptsize{\pm 0.2}}$ & $76.0 \textcolor{gray}{\scriptsize{\pm 0.2}}$ & $70.0 \textcolor{gray}{\scriptsize{\pm 0.4}}$ & $76.7 \textcolor{gray}{\scriptsize{\pm 0.4}}$ \\
& & CoTracker~\cite{karaev2023cotracker} & \ding{51} & $71.5 \textcolor{gray}{\scriptsize{\pm 0.4}}$ & $71.0 \textcolor{gray}{\scriptsize{\pm 0.3}}$ & $72.8 \textcolor{gray}{\scriptsize{\pm 0.3}}$ & $68.3 \textcolor{gray}{\scriptsize{\pm 0.7}}$ & $73.9 \textcolor{gray}{\scriptsize{\pm 0.7}}$ \\\midrule
& \multirow{2}{*}{\makecell[l]{HQ-SAM-PT\\(ours)}}
  & \multirow{2}{*}{CoTracker~\cite{karaev2023cotracker}}
  & \multirow{2}{*}{\ding{55}}
  & \multirow{2}{*}{$\mathbf{76.2} \textcolor{gray}{\scriptsize{\pm 0.1}}$} & \multirow{2}{*}{$\mathbf{75.3} \textcolor{gray}{\scriptsize{\pm 0.1}}$} & \multirow{2}{*}{$78.4 \textcolor{gray}{\scriptsize{\pm 0.2}}$} & \multirow{2}{*}{$\mathbf{72.1} \textcolor{gray}{\scriptsize{\pm 0.2}}$} & \multirow{2}{*}{$\mathbf{79.0} \textcolor{gray}{\scriptsize{\pm 0.2}}$} \\
    & & & & & & \\
\bottomrule
\end{tabular}
}
\end{table}

%% file: tables/bdd100k-vos-valid.tex
\begin{table}
\centering
\caption{This table shows the semi-supervised VOS performance on BDD100K's validation set, comparing our HQ-SAM-PT method using CoTracker as the point tracker (without reinitialization) against the zero-shot SegGPT and the fully-supervised XMem. Performance metrics include the average $\mathcal{J\&F}$ measure for object visibility durations categorized as short (1-5 frames), medium (6-30 frames), and long (31+ frames). Our approach demonstrates superior results over SegGPT for non-transient objects and over XMem across all visibility durations except for long-term object tracking.}
\vspace{-0.1in}
\label{table:vos-bdd100k-val}
\resizebox{\columnwidth}{!}{
\begin{tabular}{llcccccc}
\toprule
& & \multicolumn{6}{c}{\textbf{BDD100K VOS Validation}~\cite{bdd100k}}
\\
\cmidrule{3-8}
& \textbf{Method} & $\mathcal{J\&F}$ & $\mathcal{J}$ & $\mathcal{F}$ & \makecell[c]{$\mathcal{J\&F}$ \\ Short} & \makecell[c]{$\mathcal{J\&F}$ \\ Medium} & \makecell[c]{$\mathcal{J\&F}$ \\ Long}
\\
\toprule
\multicolumn{8}{l}{(a) trained on video segmentation data, but not on BDD100K} \\
\midrule
& XMem~\cite{cheng2022xmem}      & $76.6$ & $74.5$ & $78.7$ & $79.3$ & $78.6$ & $\mathbf{63.7}$ \\
\midrule
\multicolumn{8}{l}{(b) not trained on video segmentation data (\textit{\textbf{zero-shot}})} \\
\midrule
& SegGPT~\cite{wang2023seggpt}   & $\mathbf{81.5}$ & $\mathbf{81.2}$ & $\mathbf{81.8}$ & $\mathbf{96.1}$ & $\underline{78.6}$ & $52.0$ \\
& HQ-SAM-PT (ours)   & 
$\underline{81.0}$ & $\underline{80.1}$ & $\mathbf{81.8}$ & $\underline{91.8}$ & $\mathbf{79.9}$ & $\underline{55.8}$ \\
\bottomrule
\end{tabular}
}
\end{table}

%% file: tables/uvo-dense-video-v1.0-validation.tex
\begin{table}
\centering
\caption{{Results on the validation split of UVO~\cite{wang2021unidentified} VideoDenseSet v$\mathbf{1.0}$.} SAM-PT outperforms TAM~\cite{yang2023track} even though the former was not trained on any video segmentation data. TAM is a concurrent approach combining SAM~\cite{kirillov2023segment} and XMem~\cite{cheng2022xmem}, where XMem was pre-trained on BL30K~\cite{cheng2021modular} and trained on DAVIS~\cite{ponttuset2018davis} and YouTube-VOS~\cite{xu2018youtubevos}, but not on UVO. On the other hand, SAM-PT combines SAM with point trackers, both of which have not been trained on any video segmentation tasks.
}
\label{tab:uvo-v1-val}
\vspace{-0.1in}
\resizebox{\columnwidth}{!}{
\begin{tabular}{llccccccc}
\toprule
&
\textbf{Method}
& \textbf{Tracker}
& \textbf{Reinit}
& $\mathbf{AR100}$
& $\mathbf{ARs}$
& $\mathbf{ARm}$
& $\mathbf{ARl}$ 
& $\mathbf{AP}$ \\
\midrule
\multicolumn{9}{l}{(a) trained on video segmentation data, including UVO's training subset} \\
\midrule
& \multirow{2}{*}{\makecell[l]{\gr{Mask2Former}\\\gr{VIS~\cite{zhan2022robust}}}} & \multirow{2}{*}{\gr{-}} & \multirow{2}{*}{\gr{-}} & \multirow{2}{*}{\gr{$35.4$}} & \multirow{2}{*}{\gr{$-$}} & \multirow{2}{*}{\gr{$-$}} & \multirow{2}{*}{\gr{$-$}} & \multirow{2}{*}{\gr{$27.3$}} \\\\
& \gr{ROVIS~\cite{zhan2022robust}}           & \gr{-} & \gr{-} & \gr{$41.2$} & \gr{$-$} & \gr{$-$} & \gr{$-$} & \gr{$32.7$} \\
\midrule
\multicolumn{9}{l}{(b) trained on video segmentation data} \\
\midrule
& TAM~\cite{yang2023track}  & - & - & $24.1$ & $21.1$ & $32.9$ & $31.1$ & $1.7$ \\
\midrule
\multicolumn{9}{l}{(c) not trained on video segmentation data (\textit{\textbf{zero-shot}})} \\
\midrule
&\multirow{4}{*}{\makecell[l]{SAM-PT\\(ours)}}
 & PIPS~\cite{harley2022particle}       & \ding{55} & $28.8$ & $23.3$ & $40.8$ & $48.3$ & $\mathbf{6.7}$ \\
&& PIPS~\cite{harley2022particle}       & \ding{51} & $\mathbf{30.8}$ & $\mathbf{25.1}$ & $\mathbf{44.1}$ & $\mathbf{49.2}$ & $6.5$ \\
&& CoTracker~\cite{karaev2023cotracker} & \ding{55} & $29.5$ & $25.3$ & $39.0$ & $44.1$ & $5.8$ \\
&& CoTracker~\cite{karaev2023cotracker} & \ding{51} & $29.8$ & $25.1$ & $40.6$ & $45.6$ & $6.2$ \\
\bottomrule
\end{tabular}
}
\vspace{-0.2in}
\end{table}

%% file: tables/mose-2023-valid.tex
\begin{table}[h]
\centering
\vspace{-0.125in}
\caption{
    Quantitative results on the MOSE $2023$ validation set for semi-supervised VOS.
    Our method achieves performance comparable to the state-of-the-art zero-shot learning method. Note that SegGPT and SAM-PT adopt completely different training data.
}
\label{table:vos-mose-2023-val}
\vspace{-0.125in}
\resizebox{\columnwidth}{!}{
\begin{tabular}{llccccc}
\toprule
& \multirow{2}{*}{\textbf{Method}} & \multirow{2}{*}{\textbf{Tracker}} & \multirow{2}{*}{\textbf{Reinit}} & \multicolumn{3}{c}{\textbf{MOSE $\mathbf{2023}$ Validation}~\cite{ding2023mose}}
\\
\cmidrule{5-7}
& & & & \textbf{$\mathcal{J\&F}$ ↑} & \textbf{$\mathcal{J}$ ↑} & \textbf{$\mathcal{F}$ ↑}
\\
\toprule
\multicolumn{7}{l}{(a) trained on video segmentation data} \\
\midrule
& \gr{RDE~\cite{li2022recurrent}}      & \gr{-} & \gr{-} & \gr{$48.8$} & \gr{$44.6$} & \gr{$52.9$} \\
& \gr{XMem~\cite{cheng2022xmem}}       & \gr{-} & \gr{-} & \gr{$57.6$} & \gr{$53.3$} & \gr{$62.0$} \\
& \gr{DeAOT~\cite{yang2022decoupling}} & \gr{-} & \gr{-} & \gr{$59.4$} & \gr{$55.1$} & \gr{$63.8$} \\
& \gr{DEVA~\cite{cheng2023tracking}}   & \gr{-} & \gr{-} & \gr{$66.5$} & \gr{$62.3$} & \gr{$70.8$} \\
\midrule
\multicolumn{7}{l}{(b) not trained on video segmentation data (\textit{\textbf{zero-shot}})} \\
\midrule
& Painter~\cite{wang2023images} & - & - & $14.5$ & $10.4$ & $18.5$ \\
& SegGPT~\cite{wang2023seggpt}  & - & - & $\mathbf{45.1}$ & $\mathbf{42.2}$ & $\mathbf{48.0}$ \\
\midrule
& \multirow{5}{*}{\makecell[l]{SAM-PT\\(ours)}}
    & PIPS~\cite{harley2022particle}       & \ding{55} & $38.5 \textcolor{gray}{\scriptsize{\pm 0.2}}$ & $34.9 \textcolor{gray}{\scriptsize{\pm 0.3}}$ & $42.1 \textcolor{gray}{\scriptsize{\pm 0.2}}$ \\
& & PIPS~\cite{harley2022particle}         & \ding{51} & ${41.0} \textcolor{gray}{\scriptsize{\pm 0.5}}$ & ${38.5} \textcolor{gray}{\scriptsize{\pm 0.5}}$ & ${43.5} \textcolor{gray}{\scriptsize{\pm 0.5}}$ \\
& & CoTracker~\cite{karaev2023cotracker}   & \ding{55} & $41.8 \textcolor{gray}{\scriptsize{\pm 0.2}}$ & $38.3 \textcolor{gray}{\scriptsize{\pm 0.2}}$ & $45.2 \textcolor{gray}{\scriptsize{\pm 0.3}}$ \\
& & CoTracker~\cite{karaev2023cotracker}   & \ding{51} & $40.1 \textcolor{gray}{\scriptsize{\pm 0.5}}$ & $36.0 \textcolor{gray}{\scriptsize{\pm 0.5}}$ & $44.1 \textcolor{gray}{\scriptsize{\pm 0.4}}$ \\
& & TAPIR~\cite{doersch2023tapir}          & \ding{55} & $\underline{42.9} \textcolor{gray}{\scriptsize{\pm 0.2}}$ & $38.3 \textcolor{gray}{\scriptsize{\pm 0.2}}$ & $\underline{47.6} \textcolor{gray}{\scriptsize{\pm 0.1}}$ \\
\midrule

& \multirow{2}{*}{\makecell[l]{HQ-SAM-PT\\(ours)}}
    & \multirow{1}{*}{CoTracker~\cite{karaev2023cotracker}}
    & \multirow{1}{*}{\ding{55}}
    & \multirow{1}{*}{{$42.4$}$\textcolor{gray}{\scriptsize{\pm 0.3}}$}
    & \multirow{1}{*}{{$\underline{39.0}$}$\textcolor{gray}{\scriptsize{\pm 0.3}}$} & \multirow{1}{*}{{$45.8$}$\textcolor{gray}{\scriptsize{\pm 0.3}}$} \\
&   & TAPIR~\cite{doersch2023tapir} & \ding{55} & $42.1 \textcolor{gray}{\scriptsize{\pm 0.1}}$ & $37.6 \textcolor{gray}{\scriptsize{\pm 0.1}}$ & $46.7\textcolor{gray}{\scriptsize{\pm 0.1}}$ \\
\bottomrule
\end{tabular}
}
\vspace{-0.125in}
\end{table}

%% file: tables/davis-2016-valid.tex
\begin{table}[h]
\centering
\caption{Quantitative results on the DAVIS $2016$ validation set for semi-supervised VOS. Our method achieves higher performance compared to SegGPT, both of which are zero-shot methods.}
\label{table:vos-davis-2016-val}
\vspace{-0.125in}
\resizebox{\columnwidth}{!}{
\begin{tabular}{lccccc}
\toprule
& & & \multicolumn{3}{c}{\textbf{DAVIS 2016 Validation}~\cite{ponttuset2018davis}} \\
\cmidrule{4-6}
\textbf{Method} & \textbf{Tracker} & \textbf{Reinit} & \textbf{$\mathcal{J\&F}$ ↑} & \textbf{$\mathcal{J}$ ↑} & \textbf{$\mathcal{F}$ ↑} \\
\toprule
SegGPT~\cite{wang2023seggpt} & - & - & $82.3$ & $81.8$ & $82.8$ \\
\midrule
\multirow{4}{*}{\makecell[l]{SAM-PT}}
& PIPS~\cite{harley2022particle}       & \ding{55} & ${83.1}$\gr{\scriptsize{$\pm 1.5$}} & ${83.0}$\gr{\scriptsize{$\pm 0.8$}} & ${83.0}$\gr{\scriptsize{$\pm 1.1$}} \\
& PIPS~\cite{harley2022particle}       & \ding{51} & $80.2$\gr{\scriptsize{$\pm 0.6$}} & $80.3$\gr{\scriptsize{$\pm 0.6$}} & $80.0$\gr{\scriptsize{$\pm 0.6$}} \\
& CoTracker~\cite{karaev2023cotracker} & \ding{55} & $83.1 \pm 0.6 $ & $83.2 \pm 0.7 $ & $82.9 \pm 0.6 $ \\
& CoTracker~\cite{karaev2023cotracker} & \ding{51} & $82.6 \pm 0.8 $ & $83.0 \pm 1.0 $ & $82.2 \pm 0.9 $ \\
\midrule
HQ-SAM-PT
& CoTracker~\cite{karaev2023cotracker} & \ding{55} & $\mathbf{84.3} \pm 0.9 $ & $\mathbf{84.9} \pm 1.0 $ & $\mathbf{83.7} \pm 0.9 $ \\
\bottomrule
\end{tabular}
}
\end{table}

%% file: tables/davis-2017-testdev.tex
\begin{table}[h]
\centering
\caption{Quantitative results on the DAVIS 2017 test-dev subset for semi-supervised VOS.}
\label{table:vos-davis-2017-testdev}
\vspace{-0.125in}
\resizebox{\columnwidth}{!}{
\begin{tabular}{lccccc}
\toprule
& & & \multicolumn{3}{c}{\textbf{DAVIS 2017 Test-dev}~\cite{ponttuset2018davis}} \\
\cmidrule{4-6}
\textbf{Method} & \textbf{Tracker} & \textbf{Reinit} & \textbf{$\mathcal{J\&F}$ ↑} & \textbf{$\mathcal{J}$ ↑} & \textbf{$\mathcal{F}$ ↑} \\
\toprule
\multirow{6}{*}{\makecell[l]{SAM-PT}}
& PIPS~\cite{harley2022particle} & \ding{55} & ${62.7}$\gr{\scriptsize{$\pm 0.5$}} & ${59.4}$\gr{\scriptsize{$\pm 0.6$}} & ${66.1}$\gr{\scriptsize{$\pm 0.4$}} \\
& PIPS~\cite{harley2022particle} & \ding{51} & $61.5$\gr{\scriptsize{$\pm 1.1$}} & $59.3$\gr{\scriptsize{$\pm 1.0$}} & $63.8$\gr{\scriptsize{$\pm 1.2$}} \\
& CoTracker~\cite{karaev2023cotracker} & \ding{55} & $65.7$\gr{\scriptsize{$\pm 0.7$}} & $62.8$\gr{\scriptsize{$\pm 0.7$}} & $68.5$\gr{\scriptsize{$\pm 0.7$}} \\
& CoTracker~\cite{karaev2023cotracker} & \ding{51} & $62.0$\gr{\scriptsize{$\pm 1.2$}} & $58.8$\gr{\scriptsize{$\pm 1.2$}} & $65.1$\gr{\scriptsize{$\pm 1.1$}} \\
& TAPIR~\cite{doersch2023tapir} & \ding{55} & $\mathbf{69.0}$ & $\mathbf{66.0}$ & $\mathbf{72.0}$ \\
& TAPIR~\cite{doersch2023tapir} & \ding{51} & $64.1$ & $61.3$ & $66.9$ \\
\midrule
\multirow{1}{*}{\makecell[l]{HQ-SAM-PT}}
& \multirow{1}{*}{CoTracker~\cite{karaev2023cotracker}} & \multirow{1}{*}{\ding{55}} & \multirow{1}{*}{$64.8$\gr{\scriptsize{$\pm 0.5$}}} & \multirow{1}{*}{$61.9$\gr{\scriptsize{$\pm 0.5$}}} & \multirow{1}{*}{$67.7$\gr{\scriptsize{$\pm 0.5$}}} \\
\bottomrule
\end{tabular}
}
\end{table}

%% file: tables/ablation-sam-backbones.tex
\begin{table}[h]
    \caption{Performance of SAM-PT with different backbones and their inference speed in semi-supervised VOS on the validation subset of DAVIS $2017$, when using PIPS as the point tracker.}
    \centering
    \vspace{-0.1in}
    \resizebox{0.63\columnwidth}{!}{
        \begin{tabular}{cccr}
            \toprule
            \textbf{SAM Backbone} & $\mathcal{J\&F}$ & \textbf{FPS} \\
            \midrule
            ViT-Huge & $\mathbf{76.7}\pm0.6$ & $1.4$ \\
            ViT-Large & $76.4\pm0.6$ & $1.8$ \\
            ViT-Base & $72.2\pm0.5$ & $\mathbf{2.6}$ \\
            \bottomrule
        \end{tabular}
    }
    \label{tab:sam-backbones}
\end{table}

%% file: tables/ablation-sam-lightweight-variants.tex
\begin{table}[h]
    \caption{Performance of lightweight SAM variants and their inference speed in semi-supervised VOS on the validation subset of DAVIS $2017$ when using PIPS~\cite{harley2022particle} as the point tracker.}
    \centering
    \vspace{-0.1in}
    \resizebox{0.72\columnwidth}{!}{
        \begin{tabular}{lcccr}
            \toprule
            \textbf{SAM Variant} & \textbf{Backbone} & $\mathcal{J\&F}$ & \textbf{FPS} \\
            \midrule
            HQ-SAM~\cite{ke2023segment} & ViT-Huge & $\mathbf{77.64}$ & $1.3$ \\
            SAM~\cite{kirillov2023segment} & ViT-Huge & $76.65$ & $1.4$ \\
            Light HQ-SAM~\cite{ke2023segment} & ViT-Tiny & $71.30$ & $4.8$ \\
            MobileSAM~\cite{zhang2023faster} & ViT-Tiny & $71.07$ & $\mathbf{5.5}$ \\
            \bottomrule
        \end{tabular}
    }
    \label{tab:sam-lightweight}
\end{table}

%% file: tables/ablation-cotracker.tex
\begin{table}[h]
\centering
\caption{Ablation study on the validation subset of DAVIS $2017$ on the impact of different SAM-PT configurations using CoTracker~\cite{karaev2023cotracker} as the point tracker. PSM: point selection method. PP: positive points per mask. NP: negative points per mask. IRI: iterative refinement iterations. PS: point similarity filtering.}
\label{table:ablation-cotracker}
\vspace{-0.1in}
\resizebox{\columnwidth}{!}{
\begin{tabular}{lcccccccc}
\toprule
\multicolumn{5}{c}{\textbf{SAM-PT Config. (using CoTracker)}}
& \multicolumn{3}{c}{\textbf{DAVIS 2017 Validation}~\cite{ponttuset2018davis}} \\
\cmidrule{6-8}
{PSM} & {PP} & {NP} & {IRI} & {PS} & \textbf{$\mathcal{J\&F}$ ↑} & \textbf{$\mathcal{J}$ ↑} & \textbf{$\mathcal{F}$ ↑} \\
\toprule
\rowcolor{lightgray!16}
K-Medoids  & 16 & 1 & 12 & \ding{55} & $\mathbf{77.6}$\gr{\scriptsize{$\pm 0.7$}} & $\mathbf{74.8}$\gr{\scriptsize{$\pm 0.7$}} & $\mathbf{80.4}$\gr{\scriptsize{$\pm 0.7$}}  \\ \midrule
Shi-Tomasi & 16 & 1 & 12 & \ding{55} & $74.3$\gr{\scriptsize{$\pm 0.3$}} & $72.1$\gr{\scriptsize{$\pm 0.3$}} & $76.5$\gr{\scriptsize{$\pm 0.3$}}  \\
Random     & 16 & 1 & 12 & \ding{55} & $76.4$\gr{\scriptsize{$\pm 1.1$}} & $73.3$\gr{\scriptsize{$\pm 1.1$}} & $79.4$\gr{\scriptsize{$\pm 1.0$}}  \\
Mixed      & 16 & 1 & 12 & \ding{55} & $76.4$\gr{\scriptsize{$\pm 0.6$}} & $73.7$\gr{\scriptsize{$\pm 0.5$}} & $79.2$\gr{\scriptsize{$\pm 0.6$}}  \\ \midrule
K-Medoids  &  1 & 1 & 12 & \ding{55} & $39.0$\gr{\scriptsize{$\pm 0.9$}} & $36.1$\gr{\scriptsize{$\pm 0.9$}} & $42.0$\gr{\scriptsize{$\pm 1.0$}}  \\ \midrule
K-Medoids  & 16 & 0 & 12 & \ding{55} & $76.8$\gr{\scriptsize{$\pm 0.7$}} & $74.1$\gr{\scriptsize{$\pm 0.7$}} & $79.6$\gr{\scriptsize{$\pm 0.7$}}  \\ \midrule
K-Medoids  & 16 & 1 &  0 & \ding{55} & $75.3$\gr{\scriptsize{$\pm 0.6$}} & $73.2$\gr{\scriptsize{$\pm 0.5$}} & $77.3$\gr{\scriptsize{$\pm 0.6$}}  \\
K-Medoids  & 16 & 1 &  1 & \ding{55} & $76.6$\gr{\scriptsize{$\pm 0.6$}} & $74.3$\gr{\scriptsize{$\pm 0.6$}} & $78.9$\gr{\scriptsize{$\pm 0.6$}}  \\
K-Medoids  & 16 & 1 &100 & \ding{55} & $77.5$\gr{\scriptsize{$\pm 0.7$}} & $74.7$\gr{\scriptsize{$\pm 0.7$}} & $80.3$\gr{\scriptsize{$\pm 0.7$}}  \\ \midrule
K-Medoids  & 16 & 1 & 12 & \ding{51} & $73.8$\gr{\scriptsize{$\pm 0.8$}} & $71.1$\gr{\scriptsize{$\pm 0.8$}} & $76.5$\gr{\scriptsize{$\pm 0.8$}}  \\
\bottomrule
\end{tabular}
}
\end{table}

%% file: tables/ablation-pips-detailed.tex
\begin{table}[h]
\centering
\caption{Detailed ablation study on the validation subset of DAVIS $2017$ on the impact of different SAM-PT configurations using PIPS~\cite{harley2022particle} as the point tracker. PSM: point selection method. PP: positive points per mask. NP: negative points per mask. IRI: iterative refinement iterations. PS: point similarity filtering. RV: reinitialization variant.}
\label{table:ablation-pips-detailed}
\vspace{-0.1in}
\resizebox{\columnwidth}{!}{
\begin{tabular}{ccccccccccc}
\toprule
& \multicolumn{6}{c}{\textbf{SAM-PT Configuration (using PIPS)}} & \multicolumn{3}{c}{\textbf{DAVIS $\mathbf{2017}$ Validation}~\cite{ponttuset2018davis}} \\
\cmidrule{8-10}
& {PSM} & {PP} & {NP} & {IRI} & {PS} & {RV} & \textbf{$\mathcal{J\&F}$ ↑} & \textbf{$\mathcal{J}$ ↑} & \textbf{$\mathcal{F}$ ↑} & Gain\\
\toprule

\multicolumn{11}{l}{(a) point selection method and positive points per mask} \\
\midrule
&     Random  & $ 1$ & $0$  & $0$  & \ding{55}    & \ding{55}    & $37.1$\gr{\scriptsize{$\pm 21.7$}} & $34.3$\gr{\scriptsize{$\pm 22.0$}} & $40.0$\gr{\scriptsize{$\pm 21.5$}} & \\
&     Random  & $ 8$ & $0$  & $0$  & \ding{55}    & \ding{55}    & $70.5$\gr{\scriptsize{$\pm 1.4$}} & $68.5$\gr{\scriptsize{$\pm 1.4$}} & $72.6$\gr{\scriptsize{$\pm 1.5$}} & \\
&     Random  & $16$ & $0$  & $0$  & \ding{55}    & \ding{55}    & $70.0$\gr{\scriptsize{$\pm 1.1$}} & $68.2$\gr{\scriptsize{$\pm 1.0$}} & $71.8$\gr{\scriptsize{$\pm 1.2$}} & \\
&     Random  & $72$ & $0$  & $0$  & \ding{55}    & \ding{55}    & $62.6$\gr{\scriptsize{$\pm 0.4$}} & $62.3$\gr{\scriptsize{$\pm 0.3$}} & $62.8$\gr{\scriptsize{$\pm 0.5$}} & \\
& Shi-Tomasi  & $ 1$ & $0$  & $0$  & \ding{55}    & \ding{55}    & $20.3$\gr{\scriptsize{$\pm 0.1$}} & $18.3$\gr{\scriptsize{$\pm 0.1$}} & $22.3$\gr{\scriptsize{$\pm 0.2$}} & \\
& Shi-Tomasi  & $ 8$ & $0$  & $0$  & \ding{55}    & \ding{55}    & $72.0$\gr{\scriptsize{$\pm 0.3$}} & $70.3$\gr{\scriptsize{$\pm 0.3$}} & $73.7$\gr{\scriptsize{$\pm 0.4$}} & \\
& Shi-Tomasi  & $16$ & $0$  & $0$  & \ding{55}    & \ding{55}    & $66.6$\gr{\scriptsize{$\pm 0.4$}} & $65.7$\gr{\scriptsize{$\pm 0.5$}} & $67.6$\gr{\scriptsize{$\pm 0.4$}} & \\
& Shi-Tomasi  & $72$ & $0$  & $0$  & \ding{55}    & \ding{55}    & $54.4$\gr{\scriptsize{$\pm 0.3$}} & $54.8$\gr{\scriptsize{$\pm 0.2$}} & $54.0$\gr{\scriptsize{$\pm 0.4$}} & \\
&  K-Medoids  & $ 1$ & $0$  & $0$  & \ding{55}    & \ding{55}    & $32.2$\gr{\scriptsize{$\pm 0.7$}} & $30.4$\gr{\scriptsize{$\pm 0.8$}} & $34.0$\gr{\scriptsize{$\pm 0.7$}} & \\
\rowcolor{lightgray!16} &  K-Medoids  & $ 8$ & $0$  & $0$  & \ding{55}    & \ding{55}    & $\mathbf{72.3}$\gr{\scriptsize{$\pm 1.2$}} & $\mathbf{70.4}$\gr{\scriptsize{$\pm 1.3$}} & $\mathbf{74.3}$\gr{\scriptsize{$\pm 1.1$}} & \\
&  K-Medoids  & $16$ & $0$  & $0$  & \ding{55}    & \ding{55}    & $71.4$\gr{\scriptsize{$\pm 0.2$}} & $69.8$\gr{\scriptsize{$\pm 0.3$}} & $73.1$\gr{\scriptsize{$\pm 0.2$}} & \\
&  K-Medoids  & $72$ & $0$  & $0$  & \ding{55}    & \ding{55}    & $58.0$\gr{\scriptsize{$\pm 0.2$}} & $57.3$\gr{\scriptsize{$\pm 0.2$}} & $58.7$\gr{\scriptsize{$\pm 0.3$}} & \\
&      Mixed  & $ 1$ & $0$  & $0$  & \ding{55}    & \ding{55}    & $29.9$\gr{\scriptsize{$\pm 0.9$}} & $26.6$\gr{\scriptsize{$\pm 0.8$}} & $33.2$\gr{\scriptsize{$\pm 1.4$}} & \\
&      Mixed  & $ 8$ & $0$  & $0$  & \ding{55}    & \ding{55}    & $70.6$\gr{\scriptsize{$\pm 0.8$}} & $68.6$\gr{\scriptsize{$\pm 0.8$}} & $72.5$\gr{\scriptsize{$\pm 0.8$}} & \\
&      Mixed  & $16$ & $0$  & $0$  & \ding{55}    & \ding{55}    & $70.0$\gr{\scriptsize{$\pm 0.7$}} & $68.2$\gr{\scriptsize{$\pm 0.6$}} & $71.9$\gr{\scriptsize{$\pm 0.7$}} & \\
&      Mixed  & $72$ & $0$  & $0$  & \ding{55}    & \ding{55}    & $62.8$\gr{\scriptsize{$\pm 0.5$}} & $62.4$\gr{\scriptsize{$\pm 0.5$}} & $63.2$\gr{\scriptsize{$\pm 0.6$}} & \\
\midrule

\multicolumn{11}{l}{(b) negative points per mask} \\
\midrule
&  K-Medoids & $8$  & $0$  & $0$  & \ding{55}    & \ding{55}    & $72.3$\gr{\scriptsize{$\pm 1.2$}} & $70.4$\gr{\scriptsize{$\pm 1.3$}} & $74.3$\gr{\scriptsize{$\pm 1.1$}} & \\
\rowcolor{lightgray!16} &  K-Medoids & $8$  & $1$  & $0$  & \ding{55}    & \ding{55}    & $\mathbf{74.1}$\gr{\scriptsize{$\pm 0.7$}} & $\mathbf{72.1}$\gr{\scriptsize{$\pm 0.6$}} & $\mathbf{76.1}$\gr{\scriptsize{$\pm 0.7$}} & $\mathbf{+1.8}$ \\
&  K-Medoids & $8$  & $8$  & $0$  & \ding{55}    & \ding{55}    & $74.0$\gr{\scriptsize{$\pm 0.8$}} & $71.9$\gr{\scriptsize{$\pm 0.8$}} & $76.0$\gr{\scriptsize{$\pm 0.9$}} & \\
&  K-Medoids & $8$  & $16$ & $0$  & \ding{55}    & \ding{55}    & $73.4$\gr{\scriptsize{$\pm 0.6$}} & $71.4$\gr{\scriptsize{$\pm 0.6$}} & $75.3$\gr{\scriptsize{$\pm 0.6$}} & \\
&  K-Medoids & $8$  & $72$ & $0$  & \ding{55}    & \ding{55}    & $72.2$\gr{\scriptsize{$\pm 0.4$}} & $70.3$\gr{\scriptsize{$\pm 0.4$}} & $74.0$\gr{\scriptsize{$\pm 0.4$}} & \\
\midrule

\multicolumn{11}{l}{(c) iterative refinement iterations} \\
\midrule
&  K-Medoids & $8$  & $1$  & $0$  & \ding{55}    & \ding{55}    & $74.1$\gr{\scriptsize{$\pm 0.7$}} & $72.1$\gr{\scriptsize{$\pm 0.6$}} & $76.1$\gr{\scriptsize{$\pm 0.7$}} & \\
&  K-Medoids & $8$  & $1$  & $1$  & \ding{55}    & \ding{55}    & $75.7$\gr{\scriptsize{$\pm 0.7$}} & $73.4$\gr{\scriptsize{$\pm 0.7$}} & $78.1$\gr{\scriptsize{$\pm 0.6$}} & \\
&  K-Medoids & $8$  & $1$  & $3$  & \ding{55}    & \ding{55}    & $76.0$\gr{\scriptsize{$\pm 0.6$}} & $73.4$\gr{\scriptsize{$\pm 0.7$}} & $78.6$\gr{\scriptsize{$\pm 0.7$}} & \\
\rowcolor{lightgray!16} &  K-Medoids & $8$  & $1$  & $12$ & \ding{55}    & \ding{55}    & $\mathbf{76.3}$\gr{\scriptsize{$\pm 0.6$}} & $\mathbf{73.6}$\gr{\scriptsize{$\pm 0.6$}} & $\mathbf{78.9}$\gr{\scriptsize{$\pm 0.6$}} & $\mathbf{+2.2}$ \\
\midrule

\multicolumn{11}{l}{(d) patch similarity filtering} \\
\midrule
\rowcolor{lightgray!16} &  K-Medoids & $8$  & $1$  & $12$ & \ding{55}    & \ding{55}    & $\mathbf{76.3}$\gr{\scriptsize{$\pm 0.6$}} & $\mathbf{73.6}$\gr{\scriptsize{$\pm 0.6$}} & $\mathbf{78.9}$\gr{\scriptsize{$\pm 0.6$}} & none \\
&  K-Medoids & $8$  & $1$  & $12$ & $0.002$ & \ding{55}    & $72.7$\gr{\scriptsize{$\pm 2.0$}} & $70.2$\gr{\scriptsize{$\pm 1.8$}} & $75.2$\gr{\scriptsize{$\pm 2.1$}} & \\
&  K-Medoids & $8$  & $1$  & $12$ & $0.01$  & \ding{55}    & $70.7$\gr{\scriptsize{$\pm 2.0$}} & $68.3$\gr{\scriptsize{$\pm 1.8$}} & $73.2$\gr{\scriptsize{$\pm 2.1$}} & \\
\bottomrule

\multicolumn{11}{l}{(e) point reinitialization} \\
\midrule
&  K-Medoids & $8$  & $ 1$ & $ 0$ & \ding{55} &    A  & $75.7$\gr{\scriptsize{$\pm 0.7$}} & $73.7$\gr{\scriptsize{$\pm 0.6$}} & $77.7$\gr{\scriptsize{$\pm 0.8$}} & \\
&  K-Medoids & $8$  & $ 1$ & $ 0$ & \ding{55} &    B  & $75.8$\gr{\scriptsize{$\pm 0.6$}} & $73.5$\gr{\scriptsize{$\pm 0.9$}} & $78.1$\gr{\scriptsize{$\pm 0.3$}} & \\
&  K-Medoids & $8$  & $ 1$ & $ 0$ & \ding{55} &    C  & $75.5$\gr{\scriptsize{$\pm 0.7$}} & $73.2$\gr{\scriptsize{$\pm 0.8$}} & $77.8$\gr{\scriptsize{$\pm 0.7$}} & \\
&  K-Medoids & $8$  & $ 1$ & $ 0$ & \ding{55} &    D  & $75.4$\gr{\scriptsize{$\pm 0.2$}} & $73.3$\gr{\scriptsize{$\pm 0.2$}} & $77.5$\gr{\scriptsize{$\pm 0.3$}} & \\
&  K-Medoids & $8$  & $ 1$ & $12$ & \ding{55} &    A  & $76.6$\gr{\scriptsize{$\pm 0.8$}} & $74.0$\gr{\scriptsize{$\pm 0.8$}} & $\mathbf{79.1}$\gr{\scriptsize{$\pm 0.8$}} & \\
&  K-Medoids & $8$  & $ 1$ & $12$ & \ding{55} &    B  & $76.1$\gr{\scriptsize{$\pm 0.4$}} & $73.5$\gr{\scriptsize{$\pm 0.5$}} & $78.6$\gr{\scriptsize{$\pm 0.3$}} & \\
&  K-Medoids & $8$  & $ 1$ & $12$ & \ding{55} &    C  & $75.4$\gr{\scriptsize{$\pm 0.6$}} & $72.8$\gr{\scriptsize{$\pm 0.7$}} & $78.0$\gr{\scriptsize{$\pm 0.5$}} & \\
&  K-Medoids & $8$  & $ 1$ & $12$ & \ding{55} &    D  & $76.4$\gr{\scriptsize{$\pm 0.3$}} & $74.0$\gr{\scriptsize{$\pm 0.4$}} & $78.8$\gr{\scriptsize{$\pm 0.3$}} & \\
&  K-Medoids & $8$  & $ 1$ & $12$ & \ding{55} & \ding{55} & $76.3$\gr{\scriptsize{$\pm 0.6$}} & $73.6$\gr{\scriptsize{$\pm 0.6$}} & $78.9$\gr{\scriptsize{$\pm 0.6$}} & \\
&  K-Medoids & $8$  & $72$ & $ 0$ & \ding{55} &    A  & $74.9$\gr{\scriptsize{$\pm 0.9$}} & $73.2$\gr{\scriptsize{$\pm 0.8$}} & $76.6$\gr{\scriptsize{$\pm 1.0$}} & \\
&  K-Medoids & $8$  & $72$ & $ 0$ & \ding{55} &    B  & $76.0$\gr{\scriptsize{$\pm 1.1$}} & $73.9$\gr{\scriptsize{$\pm 1.1$}} & $78.1$\gr{\scriptsize{$\pm 1.1$}} & \\
&  K-Medoids & $8$  & $72$ & $ 0$ & \ding{55} &    C  & $75.1$\gr{\scriptsize{$\pm 0.6$}} & $72.9$\gr{\scriptsize{$\pm 0.5$}} & $77.2$\gr{\scriptsize{$\pm 0.7$}} & \\
&  K-Medoids & $8$  & $72$ & $ 0$ & \ding{55} &    D  & $75.6$\gr{\scriptsize{$\pm 1.5$}} & $73.8$\gr{\scriptsize{$\pm 1.5$}} & $77.3$\gr{\scriptsize{$\pm 1.6$}} & \\
\rowcolor{lightgray!16} &  K-Medoids & $8$  & $72$ & $12$ & \ding{55} &    A  & $\mathbf{76.8}$\gr{\scriptsize{$\pm 0.7$}} & $\mathbf{74.5}$\gr{\scriptsize{$\pm 0.8$}} & ${79.0}$\gr{\scriptsize{$\pm 0.6$}} & $\mathbf{+0.5}$ \\
&  K-Medoids & $8$  & $72$ & $12$ & \ding{55} &    B  & $74.8$\gr{\scriptsize{$\pm 0.8$}} & $72.1$\gr{\scriptsize{$\pm 0.9$}} & $77.6$\gr{\scriptsize{$\pm 0.7$}} & \\
&  K-Medoids & $8$  & $72$ & $12$ & \ding{55} &    C  & $75.0$\gr{\scriptsize{$\pm 0.4$}} & $72.1$\gr{\scriptsize{$\pm 0.4$}} & $77.8$\gr{\scriptsize{$\pm 0.5$}} & \\
&  K-Medoids & $8$  & $72$ & $12$ & \ding{55} &    D  & $75.2$\gr{\scriptsize{$\pm 1.1$}} & $72.7$\gr{\scriptsize{$\pm 1.1$}} & $77.6$\gr{\scriptsize{$\pm 1.1$}} & \\
\bottomrule

\end{tabular}
}
\end{table}

%% file: main.bbl
\begin{thebibliography}{53}
\providecommand{\natexlab}[1]{#1}
\providecommand{\url}[1]{\texttt{#1}}
\expandafter\ifx\csname urlstyle\endcsname\relax
  \providecommand{\doi}[1]{doi: #1}\else
  \providecommand{\doi}{doi: \begingroup \urlstyle{rm}\Url}\fi

\bibitem[Bay et~al.(2008)Bay, Ess, Tuytelaars, and Van~Gool]{bay2008speeded}
Herbert Bay, Andreas Ess, Tinne Tuytelaars, and Luc Van~Gool.
\newblock Speeded-up robust features (surf).
\newblock \emph{Computer vision and image understanding}, 110\penalty0
  (3):\penalty0 346--359, 2008.

\bibitem[Caelles et~al.(2018)Caelles, Montes, Maninis, Chen, {Van Gool},
  Perazzi, and Pont-Tuset]{Caelles_arXiv_2018}
Sergi Caelles, Alberto Montes, Kevis-Kokitsi Maninis, Yuhua Chen, Luc {Van
  Gool}, Federico Perazzi, and Jordi Pont-Tuset.
\newblock The 2018 davis challenge on video object segmentation.
\newblock In \emph{arXiv:1803.00557}, 2018.

\bibitem[Caelles et~al.(2019)Caelles, Pont-Tuset, Perazzi, Montes, Maninis, and
  {Van Gool}]{Caelles_arXiv_2019}
Sergi Caelles, Jordi Pont-Tuset, Federico Perazzi, Alberto Montes,
  Kevis-Kokitsi Maninis, and Luc {Van Gool}.
\newblock The 2019 davis challenge on vos: Unsupervised multi-object
  segmentation.
\newblock In \emph{arXiv:1905.00737}, 2019.

\bibitem[Caron et~al.(2021)Caron, Touvron, Misra, J{\'e}gou, Mairal,
  Bojanowski, and Joulin]{caron2021emerging}
Mathilde Caron, Hugo Touvron, Ishan Misra, Herv{\'e} J{\'e}gou, Julien Mairal,
  Piotr Bojanowski, and Armand Joulin.
\newblock Emerging properties in self-supervised vision transformers.
\newblock In \emph{ICCV}, 2021.

\bibitem[Cheng et~al.(2021{\natexlab{a}})Cheng, Choudhuri, Misra, Kirillov,
  Girdhar, and Schwing]{cheng2021mask2former}
Bowen Cheng, Anwesa Choudhuri, Ishan Misra, Alexander Kirillov, Rohit Girdhar,
  and Alexander~G. Schwing.
\newblock Mask2former for video instance segmentation.
\newblock \emph{arXiv preprint arXiv: 2112.10764}, 2021{\natexlab{a}}.

\bibitem[Cheng et~al.(2022)Cheng, Parkhi, and Kirillov]{cheng2022pointly}
Bowen Cheng, Omkar Parkhi, and Alexander Kirillov.
\newblock Pointly-supervised instance segmentation.
\newblock In \emph{CVPR}, pages 2617--2626, 2022.

\bibitem[Cheng and Schwing(2022)]{cheng2022xmem}
Ho~Kei Cheng and Alexander~G Schwing.
\newblock Xmem: Long-term video object segmentation with an atkinson-shiffrin
  memory model.
\newblock In \emph{ECCV}, 2022.

\bibitem[Cheng et~al.(2021{\natexlab{b}})Cheng, Tai, and Tang]{cheng2021mivos}
Ho~Kei Cheng, Yu-Wing Tai, and Chi-Keung Tang.
\newblock Modular interactive video object segmentation: Interaction-to-mask,
  propagation and difference-aware fusion.
\newblock In \emph{CVPR}, 2021{\natexlab{b}}.

\bibitem[Cheng et~al.(2021{\natexlab{c}})Cheng, Tai, and
  Tang]{cheng2021modular}
Ho~Kei Cheng, Yu-Wing Tai, and Chi-Keung Tang.
\newblock Modular interactive video object segmentation: Interaction-to-mask,
  propagation and difference-aware fusion.
\newblock In \emph{CVPR}, 2021{\natexlab{c}}.

\bibitem[Cheng et~al.(2023{\natexlab{a}})Cheng, Oh, Price, Schwing, and
  Lee]{cheng2023tracking}
Ho~Kei Cheng, Seoung~Wug Oh, Brian Price, Alexander Schwing, and Joon-Young
  Lee.
\newblock Tracking anything with decoupled video segmentation.
\newblock In \emph{ICCV}, 2023{\natexlab{a}}.

\bibitem[Cheng et~al.(2023{\natexlab{b}})Cheng, Li, Xu, Li, Yang, Wang, and
  Yang]{cheng2023segment}
Yangming Cheng, Liulei Li, Yuanyou Xu, Xiaodi Li, Zongxin Yang, Wenguan Wang,
  and Yi Yang.
\newblock Segment and track anything.
\newblock \emph{arXiv preprint arXiv:2305.06558}, 2023{\natexlab{b}}.

\bibitem[DeTone et~al.(2018)DeTone, Malisiewicz, and
  Rabinovich]{detone2018superpoint}
Daniel DeTone, Tomasz Malisiewicz, and Andrew Rabinovich.
\newblock Superpoint: Self-supervised interest point detection and description.
\newblock In \emph{CVPRW}, 2018.

\bibitem[Ding et~al.(2023)Ding, Liu, He, Jiang, Torr, and Bai]{ding2023mose}
Henghui Ding, Chang Liu, Shuting He, Xudong Jiang, Philip H.~S. Torr, and Song
  Bai.
\newblock Mose: A new dataset for video object segmentation in complex scenes.
\newblock \emph{arXiv preprint arXiv: 2302.01872}, 2023.

\bibitem[Doersch et~al.(2022)Doersch, Gupta, Markeeva, Recasens, Smaira, Aytar,
  Carreira, Zisserman, and Yang]{doersch2023tapvid}
Carl Doersch, Ankush Gupta, Larisa Markeeva, Adria Recasens, Lucas Smaira,
  Yusuf Aytar, Joao Carreira, Andrew Zisserman, and Yi Yang.
\newblock Tap-vid: A benchmark for tracking any point in a video.
\newblock In \emph{NeurIPS}, 2022.

\bibitem[Doersch et~al.(2023)Doersch, Yang, Vecerik, Gokay, Gupta, Aytar,
  Carreira, and Zisserman]{doersch2023tapir}
Carl Doersch, Yi Yang, Mel Vecerik, Dilara Gokay, Ankush Gupta, Yusuf Aytar,
  Joao Carreira, and Andrew Zisserman.
\newblock Tapir: Tracking any point with per-frame initialization and temporal
  refinement.
\newblock \emph{ICCV}, 2023.

\bibitem[Harley et~al.(2022)Harley, Fang, and Fragkiadaki]{harley2022particle}
Adam~W Harley, Zhaoyuan Fang, and Katerina Fragkiadaki.
\newblock Particle video revisited: Tracking through occlusions using point
  trajectories.
\newblock In \emph{ECCV}, 2022.

\bibitem[Heo et~al.(2020)Heo, Koh, and Kim]{Yuk2020IVOSGlobalLocal}
Yuk Heo, Yeong~Jun Koh, and Chang-Su Kim.
\newblock Interactive video object segmentation using global and local transfer
  modules.
\newblock In \emph{ECCV}, 2020.

\bibitem[Jabri et~al.(2020)Jabri, Owens, and Efros]{jabri2020space}
Allan Jabri, Andrew Owens, and Alexei Efros.
\newblock Space-time correspondence as a contrastive random walk.
\newblock In \emph{NeurIPS}, 2020.

\bibitem[Karaev et~al.(2023)Karaev, Rocco, Graham, Neverova, Vedaldi, and
  Rupprecht]{karaev2023cotracker}
Nikita Karaev, Ignacio Rocco, Benjamin Graham, Natalia Neverova, Andrea
  Vedaldi, and Christian Rupprecht.
\newblock Cotracker: It is better to track together.
\newblock \emph{arXiv preprint arXiv:2307.07635}, 2023.

\bibitem[Ke et~al.(2023)Ke, Ye, Danelljan, Liu, Tai, Tang, and
  Yu]{ke2023segment}
Lei Ke, Mingqiao Ye, Martin Danelljan, Yifan Liu, Yu-Wing Tai, Chi-Keung Tang,
  and Fisher Yu.
\newblock Segment anything in high quality.
\newblock In \emph{NeurIPS}, 2023.

\bibitem[Kirillov et~al.(2023)Kirillov, Mintun, Ravi, Mao, Rolland, Gustafson,
  Xiao, Whitehead, Berg, Lo, Dollar, and Girshick]{kirillov2023segment}
Alexander Kirillov, Eric Mintun, Nikhila Ravi, Hanzi Mao, Chloe Rolland, Laura
  Gustafson, Tete Xiao, Spencer Whitehead, Alexander~C. Berg, Wan-Yen Lo, Piotr
  Dollar, and Ross Girshick.
\newblock Segment anything.
\newblock In \emph{ICCV}, pages 4015--4026, 2023.

\bibitem[Lampert et~al.(2009)Lampert, Nickisch, and
  Harmeling]{lampert2009learning}
Christoph~H Lampert, Hannes Nickisch, and Stefan Harmeling.
\newblock Learning to detect unseen object classes by between-class attribute
  transfer.
\newblock In \emph{CVPR}, pages 951--958. IEEE, 2009.

\bibitem[Li et~al.(2022)Li, Hu, Xiong, Zhang, Pan, and Liu]{li2022recurrent}
Mingxing Li, Li Hu, Zhiwei Xiong, Bang Zhang, Pan Pan, and Dong Liu.
\newblock Recurrent dynamic embedding for video object segmentation.
\newblock In \emph{CVPR}, 2022.

\bibitem[Lin et~al.(2021)Lin, Xie, Li, and Zhang]{lin2021query}
Fanchao Lin, Hongtao Xie, Yan Li, and Yongdong Zhang.
\newblock Query-memory re-aggregation for weakly-supervised video object
  segmentation.
\newblock In \emph{AAAI}, 2021.

\bibitem[Lowe(2004)]{lowe2004distinctive}
David~G Lowe.
\newblock Distinctive image features from scale-invariant keypoints.
\newblock \emph{IJCV}, 60:\penalty0 91--110, 2004.

\bibitem[Lucas and Kanade(1981)]{lucas1981iterative}
Bruce~D Lucas and Takeo Kanade.
\newblock An iterative image registration technique with an application to
  stereo vision.
\newblock In \emph{IJCAI}, 1981.

\bibitem[Miao et~al.(2020)Miao, Wei, and
  Yang]{miao2020memoryAggregationInteractive}
Jiaxu Miao, Yunchao Wei, and Yi Yang.
\newblock Memory aggregation networks for efficient interactive video object
  segmentation.
\newblock In \emph{CVPR}, 2020.

\bibitem[Oh et~al.(2019)Oh, Lee, Xu, and Kim]{oh2019fastInteractive}
Seoung~Wug Oh, Joon-Young Lee, Ning Xu, and Seon~Joo Kim.
\newblock Fast user-guided video object segmentation by
  interaction-and-propagation networks.
\newblock In \emph{CVPR}, 2019.

\bibitem[Park and Jun(2009)]{park2009kmedoids}
Hae-Sang Park and Chi-Hyuck Jun.
\newblock A simple and fast algorithm for k-medoids clustering.
\newblock \emph{Expert Systems with Applications}, 36\penalty0 (2, Part
  2):\penalty0 3336--3341, 2009.

\bibitem[Pont-Tuset et~al.(2017)Pont-Tuset, Perazzi, Caelles, Arbel\'aez,
  Sorkine-Hornung, and {Van Gool}]{ponttuset2018davis}
Jordi Pont-Tuset, Federico Perazzi, Sergi Caelles, Pablo Arbel\'aez, Alexander
  Sorkine-Hornung, and Luc {Van Gool}.
\newblock The 2017 davis challenge on video object segmentation.
\newblock \emph{arXiv:1704.00675}, 2017.

\bibitem[Radford et~al.(2021)Radford, Kim, Hallacy, Ramesh, Goh, Agarwal,
  Sastry, Askell, Mishkin, Clark, et~al.]{radford2021learning}
Alec Radford, Jong~Wook Kim, Chris Hallacy, Aditya Ramesh, Gabriel Goh,
  Sandhini Agarwal, Girish Sastry, Amanda Askell, Pamela Mishkin, Jack Clark,
  et~al.
\newblock Learning transferable visual models from natural language
  supervision.
\newblock 2021.

\bibitem[Sand and Teller(2008)]{sand2008particle}
Peter Sand and Seth Teller.
\newblock Particle video: Long-range motion estimation using point
  trajectories.
\newblock \emph{IJCV}, 80:\penalty0 72--91, 2008.

\bibitem[Sarlin et~al.(2020)Sarlin, DeTone, Malisiewicz, and
  Rabinovich]{sarlin2020superglue}
Paul-Edouard Sarlin, Daniel DeTone, Tomasz Malisiewicz, and Andrew Rabinovich.
\newblock Superglue: Learning feature matching with graph neural networks.
\newblock In \emph{CVPR}, 2020.

\bibitem[Shi and Tomasi(1994)]{shi1994good}
Jianbo Shi and Tomasi.
\newblock Good features to track.
\newblock In \emph{CVPR}, 1994.

\bibitem[Teed and Deng(2020)]{teed2020raft}
Zachary Teed and Jia Deng.
\newblock Raft: Recurrent all-pairs field transforms for optical flow.
\newblock In \emph{ECCV}, 2020.

\bibitem[Tomasi and Kanade(1991)]{tomasi1991detection}
Carlo Tomasi and Takeo Kanade.
\newblock Detection and tracking of point.
\newblock \emph{IJCV}, 9:\penalty0 137--154, 1991.

\bibitem[Wang et~al.(2005)Wang, Bhat, Colburn, Agrawala, and
  Cohen]{wang2005interactiveVideoCutout}
Jue Wang, Pravin Bhat, R~Alex Colburn, Maneesh Agrawala, and Michael~F Cohen.
\newblock Interactive video cutout.
\newblock In \emph{ToG}, 2005.

\bibitem[Wang et~al.(2019)Wang, Zhang, Bertinetto, Hu, and Torr]{wang2019fast}
Qiang Wang, Li Zhang, Luca Bertinetto, Weiming Hu, and Philip~HS Torr.
\newblock Fast online object tracking and segmentation: A unifying approach.
\newblock In \emph{CVPR}, 2019.

\bibitem[Wang et~al.(2023{\natexlab{a}})Wang, Chang, Cai, Li, Hariharan,
  Holynski, and Snavely]{wang2023omnimotion}
Qianqian Wang, Yen-Yu Chang, Ruojin Cai, Zhengqi Li, Bharath Hariharan,
  Aleksander Holynski, and Noah Snavely.
\newblock Tracking everything everywhere all at once.
\newblock In \emph{ICCV}, 2023{\natexlab{a}}.

\bibitem[Wang et~al.(2021)Wang, Feiszli, Wang, and Tran]{wang2021unidentified}
Weiyao Wang, Matt Feiszli, Heng Wang, and Du Tran.
\newblock Unidentified video objects: A benchmark for dense, open-world
  segmentation.
\newblock In \emph{ICCV}, 2021.

\bibitem[Wang et~al.(2023{\natexlab{b}})Wang, Wang, Cao, Shen, and
  Huang]{wang2023images}
Xinlong Wang, Wen Wang, Yue Cao, Chunhua Shen, and Tiejun Huang.
\newblock Images speak in images: A generalist painter for in-context visual
  learning.
\newblock In \emph{CVPR}, 2023{\natexlab{b}}.

\bibitem[Wang et~al.(2023{\natexlab{c}})Wang, Zhang, Cao, Wang, Shen, and
  Huang]{wang2023seggpt}
Xinlong Wang, Xiaosong Zhang, Yue Cao, Wen Wang, Chunhua Shen, and Tiejun
  Huang.
\newblock Seggpt: Segmenting everything in context.
\newblock In \emph{ICCV}, 2023{\natexlab{c}}.

\bibitem[Wu et~al.(2022)Wu, Jiang, Zhang, Bai, and Bai]{wu2021seqformer}
Junfeng Wu, Yi Jiang, Wenqing Zhang, Xiang Bai, and Song Bai.
\newblock Seqformer: a frustratingly simple model for video instance
  segmentation.
\newblock In \emph{ECCV}, 2022.

\bibitem[Xu et~al.(2018)Xu, Yang, Fan, Yue, Liang, Yang, and
  Huang]{xu2018youtubevos}
Ning Xu, Linjie Yang, Yuchen Fan, Dingcheng Yue, Yuchen Liang, Jianchao Yang,
  and Thomas Huang.
\newblock Youtube-vos: A large-scale video object segmentation benchmark, 2018.

\bibitem[Yang et~al.(2021)Yang, Lamdouar, Lu, Zisserman, and Xie]{yang2021self}
Charig Yang, Hala Lamdouar, Erika Lu, Andrew Zisserman, and Weidi Xie.
\newblock Self-supervised video object segmentation by motion grouping.
\newblock In \emph{ICCV}, 2021.

\bibitem[Yang et~al.(2023)Yang, Gao, Li, Gao, Wang, and Zheng]{yang2023track}
Jinyu Yang, Mingqi Gao, Zhe Li, Shang Gao, Fangjing Wang, and Feng Zheng.
\newblock Track anything: Segment anything meets videos.
\newblock \emph{arXiv preprint arXiv:2304.11968}, 2023.

\bibitem[Yang and Yang(2022{\natexlab{a}})]{yang2022deaot}
Zongxin Yang and Yi Yang.
\newblock Decoupling features in hierarchical propagation for video object
  segmentation.
\newblock In \emph{NeurIPS}, 2022{\natexlab{a}}.

\bibitem[Yang and Yang(2022{\natexlab{b}})]{yang2022decoupling}
Zongxin Yang and Yi Yang.
\newblock Decoupling features in hierarchical propagation for video object
  segmentation.
\newblock In \emph{NeurIPS}, 2022{\natexlab{b}}.

\bibitem[Yi et~al.(2016)Yi, Trulls, Lepetit, and Fua]{yi2016lift}
K.~M. Yi, Eduard Trulls, Vincent Lepetit, and P. Fua.
\newblock Lift: Learned invariant feature transform.
\newblock \emph{ECCV}, 2016.

\bibitem[Yu et~al.(2020)Yu, Chen, Wang, Xian, Chen, Liu, Madhavan, and
  Darrell]{bdd100k}
Fisher Yu, Haofeng Chen, Xin Wang, Wenqi Xian, Yingying Chen, Fangchen Liu,
  Vashisht Madhavan, and Trevor Darrell.
\newblock Bdd100k: A diverse driving dataset for heterogeneous multitask
  learning.
\newblock In \emph{IEEE/CVF Conference on Computer Vision and Pattern
  Recognition (CVPR)}, 2020.

\bibitem[Zhan et~al.(2022)Zhan, McKee, and Lazebnik]{zhan2022robust}
Zitong Zhan, Daniel McKee, and Svetlana Lazebnik.
\newblock Robust online video instance segmentation with track queries.
\newblock \emph{arXiv preprint arXiv: 2211.09108}, 2022.

\bibitem[Zhang et~al.(2023)Zhang, Han, Qiao, Kim, Bae, Lee, and
  Hong]{zhang2023faster}
Chaoning Zhang, Dongshen Han, Yu Qiao, Jung~Uk Kim, Sung-Ho Bae, Seungkyu Lee,
  and Choong~Seon Hong.
\newblock Faster segment anything: Towards lightweight sam for mobile
  applications.
\newblock \emph{arXiv preprint arXiv:2306.14289}, 2023.

\bibitem[Zheng et~al.(2023)Zheng, Harley, Shen, Wetzstein, and
  Guibas]{zheng2023point}
Yang Zheng, Adam~W. Harley, Bokui Shen, Gordon Wetzstein, and Leonidas~J.
  Guibas.
\newblock Pointodyssey: A large-scale synthetic dataset for long-term point
  tracking.
\newblock In \emph{ICCV}, 2023.

\end{thebibliography}
